\def\year{2022}\relax
\documentclass[letterpaper]{article} 
\usepackage{aaai22}  
\usepackage{times}  
\usepackage{helvet}  
\usepackage{courier}  
\usepackage[hyphens]{url}  
\usepackage{graphicx} 
\usepackage{natbib}  
\usepackage{caption} 
\DeclareCaptionStyle{ruled}{labelfont=normalfont,labelsep=colon,strut=off} 
\usepackage{algorithm}
\usepackage{algorithmic}
\usepackage{spreadtab}
\usepackage{amsmath}
\usepackage{subcaption}
\usepackage{booktabs}
\usepackage{color}
\usepackage{amssymb}

\usepackage{newfloat}
\usepackage{listings}
\floatstyle{ruled}
\newfloat{listing}{tb}{lst}{}
\floatname{listing}{Listing}
\title{SCALoss: Side and Corner Aligned Loss for Bounding Box Regression}
\author{
    Tu Zheng\textsuperscript{\rm 1,2},
    Shuai Zhao\textsuperscript{\rm 1},
    Yang Liu\textsuperscript{\rm 1},
    Zili Liu\textsuperscript{\rm 1,2},
    Deng Cai\textsuperscript{\rm 1,2}\thanks{Deng Cai is the corresponding author.}\\
}
\affiliations{
    \textsuperscript{\rm 1}State Key Lab of CAD\&CG, Zhejiang University, China  \\
    \textsuperscript{\rm 2}Fabu Inc., Hangzhou, China \\
   { \{zhengtuzju,zhaoshuaimcc,zililiuzju\}@gmail.com, \{lyng\_95, dcai\}@zju.edu.cn}  \\

}

\usepackage{bibentry}

\begin{document}

\maketitle

\begin{abstract}
Bounding box regression is an important component in object detection. Recent work achieves promising performance by optimizing the Intersection over Union~(IoU). However, IoU-based loss has the gradient vanish problem in the case of low overlapping bounding boxes, and the model could easily ignore these simple cases. In this paper, we propose Side Overlap~(SO) loss by maximizing the side overlap of two bounding boxes, which puts more penalty for low overlapping bounding box cases. Besides, to speed up the convergence, the Corner Distance~(CD) is added into the objective function. Combining the Side Overlap and Corner Distance, we get a new regression objective function, \textit{Side and Corner Align Loss~(SCALoss)}. The SCALoss is well-correlated with IoU loss, which also benefits the evaluation metric but produces more penalty for low-overlapping cases. It can serve as a comprehensive similarity measure, leading to better localization performance and faster convergence speed. Experiments on COCO, PASCAL VOC, and LVIS benchmarks show that SCALoss can bring consistent improvement and outperform $\ell_n$ loss and IoU based loss with popular object detectors such as YOLOV3, SSD, Faster-RCNN. Code is available at: \url{https://github.com/Turoad/SCALoss}.
\end{abstract}

\section{Introduction}

Object detection has been improved rapidly with the development of advanced deep convolutional neural networks. A series of state-of-the-art CNN-based detectors emerge in recent years, such as Faster R-CNN~\cite{ren2015faster}, SSD~\cite{liu2016ssd}, YOLOV3~\cite{redmon2018yolov3}, Reppoints~\cite{yang2019reppoints}, and \emph{etc}. Generally, object detection consists of object classification and object localization. Current state-of-the-art object detectors~(\emph{e.g.} Faster-RCNN, Mask R-CNN~\cite{he2017mask}, RetinaNet~\cite{lin2017focal}) have shown the importance of bounding box regression in object detection pipeline. In this paper, we focus on the problem of object localization. 

Intersection over Union~(IoU) is the most popular evaluation metric for bounding box regression. In the existing methods, $\ell_n$ loss is the widely used loss, but it is not tailored to the evaluation metric~(IoU). Thus, IoU loss~\cite{yu2016unitbox} is proposed to directly optimize the evaluation metric. However, IoU is infeasible to optimize in the case of non-overlapping bounding boxes. Then Generalized IoU~(GIoU) loss ~\cite{rezatofighi2019generalized} addresses this weakness by introducing a generalized version as the new loss. After that, Distance IoU~(DIoU) loss~\cite{zheng2019distance} adds the normalized center distance between the predicted box and the target box, which helps converge faster than GIoU loss. Although the IoU-based loss can achieve more accurate result than $\ell_n$ loss, they still have several limitations as shown in Fig.~\ref{fig1}. Below, we describe these issues in turn:

\begin{figure}[!t]
  \centering
  \includegraphics[width=0.49\columnwidth]{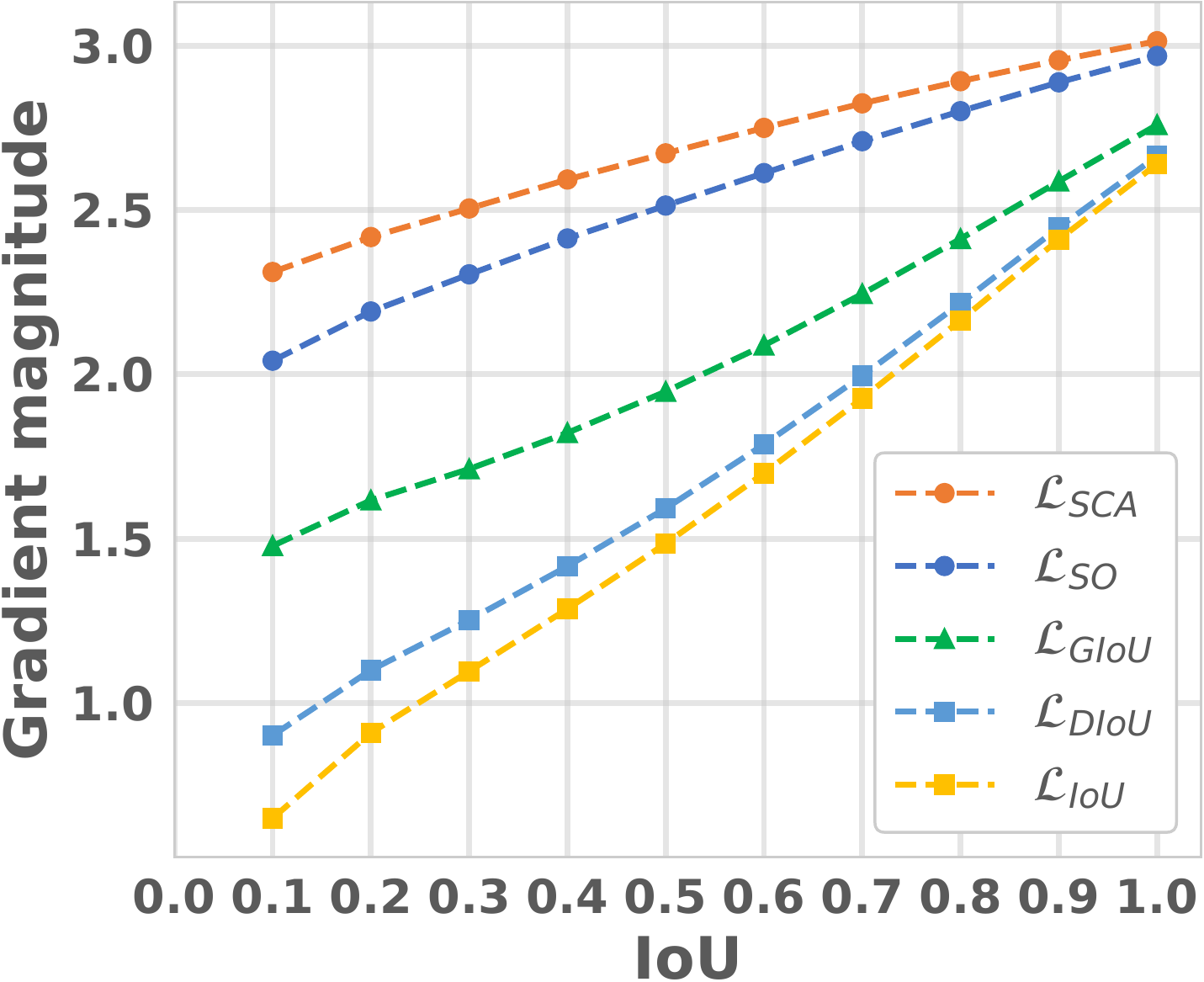}
  \includegraphics[width=0.49\columnwidth]{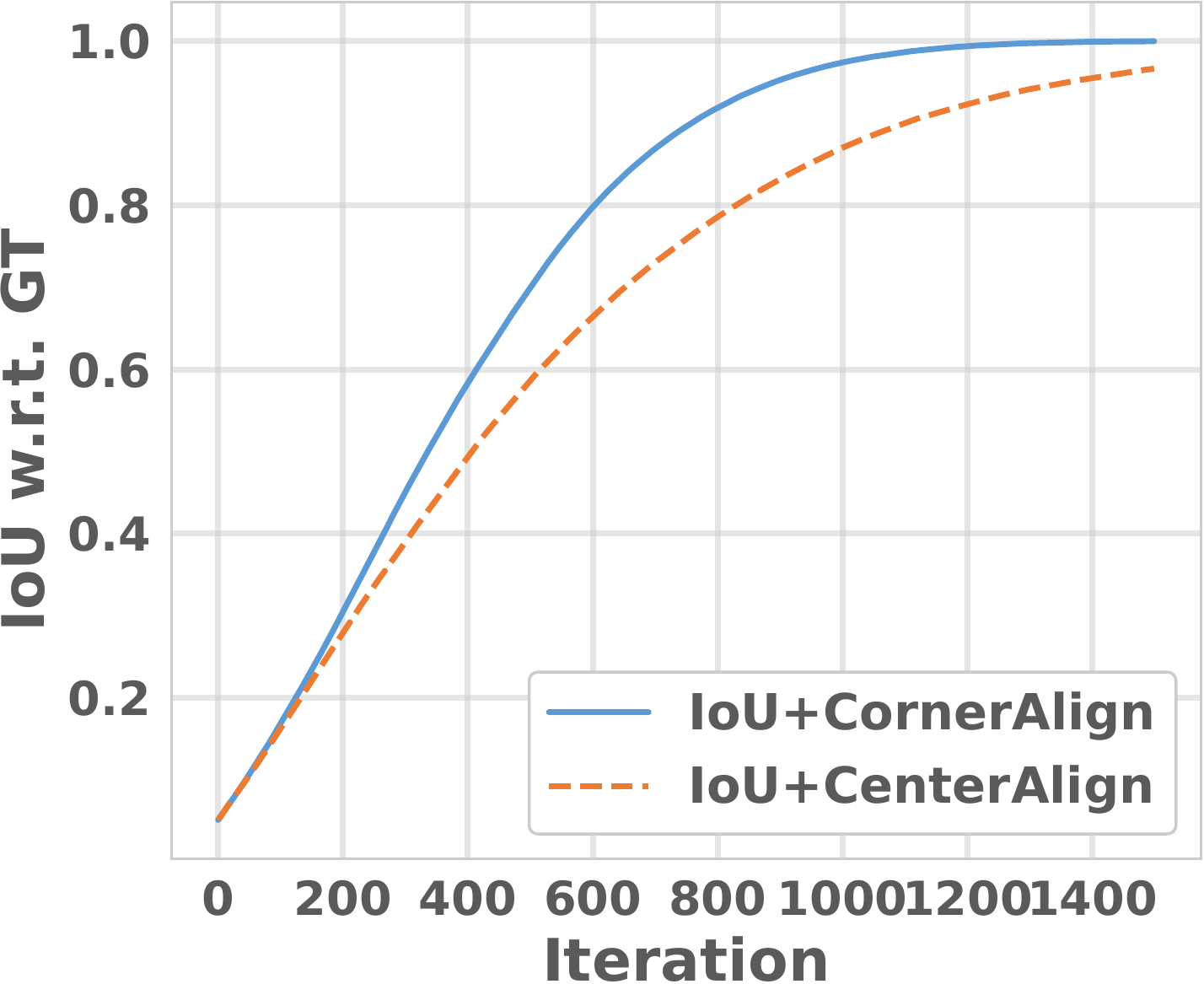}
\caption{(left)~The relationship between gradient magnitude and IoU. Gradient magnitude is the mean of $ || \frac{\partial{\mathcal{L}} } {\partial{x}}||_2$ in different IoU intervals, where IoU interval = \{[0., 0.1], [0.1, 0.2], ..., [0.9, 1.0]\}, $x = (x_1, y_1, x_2, y_2)$ is the bounding box corner point. The gradient of $\mathcal{L}_{IoU}$  significantly drops in low overlapping cases, but $\mathcal{L}_{SO}$, $\mathcal{L}_{SCA}$ still have a large gradient. All results are calculated with 1000k random boxes. (right)~A comparison between center alignment and corner alignment. The loss weight of corner alignment is 0.5x as center alignment. Detail settings can be found in the Approach section.}
  \label{fig1}
\end{figure}

\begin{figure}[!t]
\centering
  \begin{subfigure}{0.15\textwidth}
    \centering
    \includegraphics[width=0.99\linewidth]{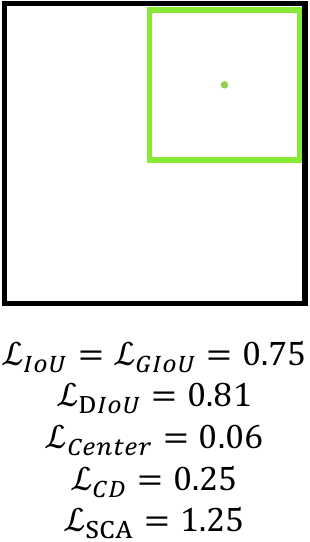}
  \end{subfigure}
  \begin{subfigure}{0.15\textwidth}
    \centering
    \includegraphics[width=0.99\linewidth]{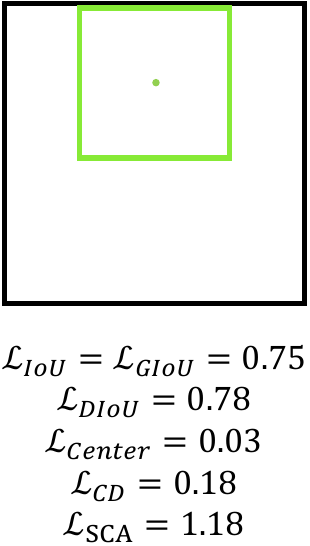}
  \end{subfigure}
  \begin{subfigure}{0.15\textwidth}
    \centering
    \includegraphics[width=0.99\linewidth]{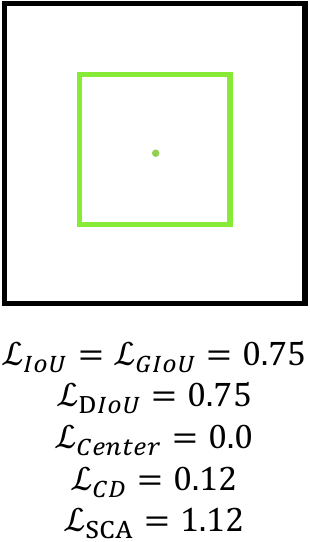}
  \end{subfigure}
\caption{
Green and black bounding box denote the predicted box and ground truth box respectively. $\mathcal{L}_{CD}$ is Eq.\eqref{loss_cd} and $\mathcal{L}_{Center}$ is the normalized center distance in DIoU. In thest cases, GIoU degrades into IoU and DIoU heavily relies IoU yielding inferior performance.
}
\label{iou_shortcommoning}
\end{figure}

\textbf{1) Gradient vanish problem}: IoU-based methods~(IoU, GIoU, DIoU) improve baseline for high overlapping metric like AP75, but have relatively inferior performance in AP50. We further investigate this phenomenon and notice that IoU loss will lead to gradient vanish problem for non-overlapping cases and produce small gradient for low overlapping cases. In Fig.~\ref{fig1}~(left), we visualize the relationship between gradient magnitude and IoU for different loss functions. It shows that lower IoU cases will have relatively smaller gradient value. During the training process, the small gradients produced by low overlapping boxes~(hard samples) may be drowned into the large gradients by high overlapping ones~(easy samples), thus limiting the overall performance. Since IoU is a component of GIoU and DIoU, they still encounter this problem. When predicted boxes lie within ground truth boxes, as shown in Fig. \ref{iou_shortcommoning}, all IoU value is the same and GIoU degrades into IoU. When the center of box is close to its ground truth, the normalized center distance in DIoU is near zero. In this case, the DIoU is roughly the same as IoU. For the aforementioned cases, GIoU and DIoU still produce small gradient, resulting in inferior performance. 

\textbf{2) Slow convergence speed}: Although DIoU~\cite{zheng2019distance} can speed up the convergence to a certain, the designed objective function is still not optimal. As DIoU discusses, GIoU tends to increase the size of box for non-overlapping cases until it has overlap with the ground truth box, which makes GIoU slow for convergence. Thus, DIoU adds a penalty term, \emph{i.e.}, the normalized center distance to directly ``pull'' closer boxes, which makes the DIoU converge faster than GIoU. However, as Fig.~\ref{iou_shortcommoning} shows, DIoU contributes little when the center of predicted box is near target box. In these cases, the corner distance is still far from the ground truth box. We further compare center alignment~(regressing normalized center distance) with the corner alignment~(regressing normalized corner distance) as shown in Fig.~\ref{fig1}~(right). It illustrates the corner alignment converges faster than center alignment. Therefore, regressing the two corner points can be a better choice.

In this work, we propose \textit{Side and Corner Aligned Loss (SCALoss)} to solve the shortcoming of IoUs and speed up the convergence. It is a combination of \textit{Side Overlap~(SO) loss}~($\mathcal{L}_{SO}$, Eq.~\eqref{loss_so}) and \textit{Corner Distance~(CD) loss}~($\mathcal{L}_{LD}$, Eq.~\eqref{loss_cd}). \emph{The Side Overlap maximizes the side overlap of bounding boxes, which puts more penalty for low-overlapping cases and focuses more on hard samples.} As shown in Fig.~\ref{fig1}~(left), SO still keeps a large gradient in the low overlapping cases, while the gradient of IoU significantly drops. Furthermore, SO loss is well-correlated with IoU loss~(see the Sec.~\textbf{Relationship with IoU and GIoU}). Specifically, it can also benefit the evaluation metric~(IoU). \emph{The Corner Distance adds the normalized corner distance to achieve accurate corner alignment and faster convergence speed.} By incorporating the Side Overlap loss and Corner Distance loss, SCALoss can serve a more comprehensive similarity measure, leading the better localization performance and faster convergence speed. 

To demonstrate the generality of SCALoss, we evaluate it with various CNN-based object detection frameworks including YOLOV3~\cite{redmon2018yolov3}, SSD~\cite{liu2016ssd}, Faster R-CNN~\cite{ren2015faster} on PASCAL VOC~\cite{everingham2010pascal}, MS-COCO~\cite{lin2014microsoft} , and LVIS~\cite{gupta2019lvis} datesets. Experimental results demonstrate that our approach achieves better object localization accuracy and gets consistent improvements.

Our contributions can be summarized as follows:
\begin{itemize}
    \item We show that IoU based method has the gradient problem for low overlapping bounding boxes and the normalized corner distance can speed up convergence. 
    \item We propose SCALoss to evaluate the similarity of bounding boxes by using corner points and box sides, which outperforms $\ell_n$ loss and IoU-based loss~(including IoU~\cite{yu2016unitbox}, GIoU~\cite{rezatofighi2019generalized},  DIoU~\cite{zheng2019distance}, and CIoU~\cite{zheng2019distance}). It can be easily plugged into any detection framework to achieve better localization accuracy.
    \item We experimentally demonstrate that SCALoss can achieve noticeable and consistent improvement with different detection frameworks on PASCAL VOC, COCO, and LVIS benchmarks.
\end{itemize}{}

\section{Related Work}

\subsection{Object Detection}

Current object detection methods can be roughly categorized into two classes: anchor-based detectors and anchor-free detectors. Anchor-based detectors can be divided into two-stage and one-stage methods.

\subsubsection{Anchor-based Detectors}
Anchor-based detectors consist of two-stage detectors and one-stage detectors.
For two-stage detectors, R-CNN based methods~\cite{girshick2014rich, girshick2015fast, ren2015faster} generate object proposals with sliding window for second stage classifier as well as bounding box refinement. After that, lots of algorithms are proposed to improve its performance~\cite{dai2016r, cai2018cascade, shrivastava2016contextual,li2019scale,lu2019grid}. Compared to two-stage methods, the one-stage detectors directly predict bounding boxes and class scores without object proposal generation such as SSD~\cite{liu2016ssd} and YOLO series~\cite{redmon2016you,redmon2017yolo9000,redmon2018yolov3, bochkovskiy2020yolov4, wang2020scaled}. Thereafter, plenty of works are presented to boost its performance\cite{fu2017dssd, kong2017ron, zhang2018single}. These methods are superior in inference speed but inferior in accuracy compared to two-stage methods. Among these methods, Focal loss~\cite{lin2017focal} solves the problem of extreme foreground-background class imbalance.  Generally, one-stage method is considered to be promising to achieve similar accuracy with two-stage method.

\subsubsection{Anchor-free Detectors} Anchor-free detectors mainly locate several pre-defined keypoints and generate bounding boxes to detect objects. CornerNet~\cite{law2018cornernet} detects an object bounding box as a pair of keypoints while CenterNet~\cite{duan2019centernet} detects object center and regress the size of the object. ExtremeNet~\cite{zhou2019bottom} detects four extreme points and one center to generate the object bounding box.
Reppoints~\cite{yang2019reppoints} represents objects as a set of sample points to adaptively position themselves over an object and utilizes deformable convolution~\cite{zhu2019deformable} to get more accurate features. These anchor-free detectors are able to eliminate those hyper-parameters related to anchors and have achieved similar performance with anchor-based detectors.

\subsection{Bounding Box Regression Loss}

Various bounding box regression losses have been proposed in recent years.
$\ell_1$-smooth loss~\cite{girshick2015fast} proposes to combine $\ell_1$ loss and $\ell_2$ loss so that the loss is less sensitive to outliers and more stable for inliers. Balanced L1 loss~\cite{pang2019libra} proposes to promote the crucial regression gradients~(inliers) for the balance between classification and localization. Bounded IoU loss~\cite{tychsen2018improving} derives a novel bounding
box regression loss based on a set of IoU upper bounds that better matches the goal of IoU maximization while still providing good convergence properties. KLLoss~\cite{he2019bounding} proposes a bounding box regression loss for learning bounding box transformation and localization variance together. The learned localization variance can merge neighboring bounding boxes during non-maximum suppression (NMS), which further improves the localization performance. UnitBox~\cite{yu2016unitbox} first proposes IoU Loss for object detection, which regresses the bounding box as a whole unit. GIoU~\cite{rezatofighi2019generalized} discusses the weakness of IoU for the case of non-overlapping bounding boxes and introduces a generalized version of IoU as a new loss. DIoU~\cite{zheng2019distance} adds the normalized center distance between the predicted and the target box on IoU loss, which helps converge faster in training. CIoU~\cite{zheng2019distance} suggests  the consistency of aspect ratios for bounding boxes is also an important geometric factor and extends DIoU by regressing aspect ratios, leading to better performance.

\section{Approach}

In this section, we first introduce our Side and Corner Aligned loss for bounding box regression, then we analyze the SO loss and compare it with IoU based loss.

\subsection{Side and Corner Aligned Loss}
Following~\cite{he2019bounding}, we regress the corners of a bounding box separately. We adopt the parameter of the $(x_1, y_1,x_2,y_2) \in \mathcal{R}^4$ coordinate as bounding box representation, where $(x_1, y_1)$, $(x_2,y_2)$ are top left and bottom  right corner respectively.  Our loss function includes side overlap~(SO) and corner distance~(CD) two parts.

\begin{figure}[!t]
\centering
    \begin{subfigure}{0.22\textwidth}
    \includegraphics[width=0.99\columnwidth]{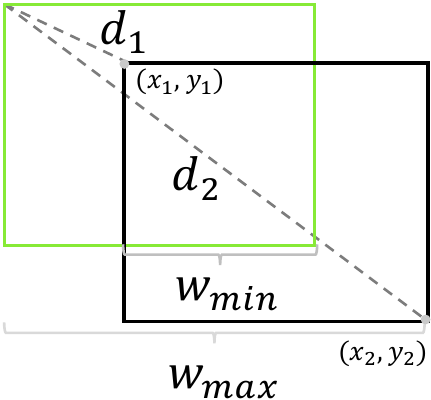}
    \caption{Overlapping case}
    \label{overlapping_loss}
    \end{subfigure}
    \begin{subfigure}{0.22\textwidth}
    \includegraphics[width=0.99\columnwidth]{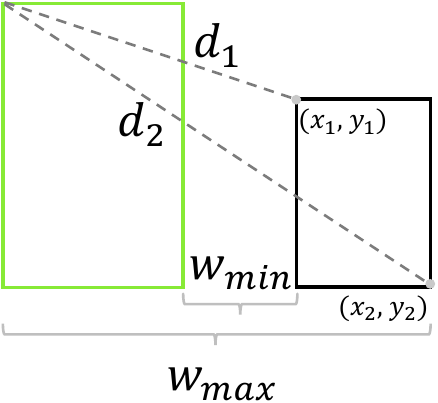}
    \caption{Non-overlapping case}
    \label{non_overlapping_loss}
    \end{subfigure}
\caption{
Side and Corner Align loss for the bounding box regression.
It directly regresses two corner points by minimizing the normalized distance $\frac{d_1 }{d_2}$ and enlarges width by minimizing the $1-\frac{w_{min}}{w_{max}}$. We omit the another corner and side for simplicity.
}
\label{regression loss}
\end{figure}

\subsubsection{Side Overlap}
We propose Side Overlap (SO) loss to measure bounding box similarity by maximizing the overlap of width and height. It is a stricter constraint and puts more gradient for low overlapping bounding box. As shown in Fig. \ref{regression loss}, given the predicted box~($x_1,x_2, y_1,y_2$) and ground truth box~($x_1^g,y_1^g,x_2^g,y_2^g$), the SO loss simultaneously maximizes the overlap for both sides of a predicted box with its ground truth. SO is defined as follows:
\begin{align}
SO = \frac{w_{min}}{w_{max}} + \frac{h_{min}}{h_{max}},
\end{align}{}where $w_{min} = \min(x_2, x_2^g) - \max(x_1, x_1^g)$, $w_{max}=\max(x_2, x_2^g) - \min(x_1, x_1^g)$, $h_{min} = \min(y_2, y_2^g) - \max(y_1, y_1^g)$, $h_{max}=\max(y_2, y_2^g) - \min(y_1, y_1^g)$. 
Note $w_{min}$, $h_{min}$ may be negative when bounding boxes are non-overlapping as shown in Fig. \ref{non_overlapping_loss}. Thus, SO loss can also be optimized for non-overlapping cases. The SO loss can be formulated as follows:
\begin{align}  \label{loss_so}
    \mathcal{L}_{SO} = 2 - SO.
\end{align}{}

\subsubsection{Corner Distance}
Furthermore, we introduce Corner Distance (CD) loss to achieve better corner alignment. As shown in Fig. \ref{iou_shortcommoning}, the normalized center distance in DIoU is roughly near zero in these cases, but the corner point still misaligns. Therefore, we add CD in the loss to achieve accurate box regression. The CD directly minimizes the normalized corner distance. It is defined as follows:
\begin{align} \label{loss_cd}
    \mathcal{L}_{CD} = \frac{D(p_1, p_1^g)}{D(p_{c_1}, p_{c_2})}+
    \frac{D(p_2, p_2^g)}{D(p_{c_1}, p_{c_2})},
\end{align}{}where $D(\cdot, \cdot)$ is the Euclidean distance, $p_1$, $p_2$ denote the top left and bottom right corner points of predicted box, $p_1^g$, $p_2^g$ are corresponding ground truth points, $p_{c_1}$, $p_{c_2}$ are corner points of the smallest enclosing box covering two boxes.

The final SCALoss can be formulated as follows:
\begin{align} \label{final_loss}
\mathcal{L}_{SCA} = \mathcal{L}_{SO} + \alpha \mathcal{L}_{CD},
\end{align}{}where $\alpha$ is the weight factor, and $\alpha$ is set to 0.5 in
our experiment.

\begin{figure}[!t]
\centering
\begin{subfigure}{0.49\columnwidth}
\includegraphics[width=0.99\linewidth]{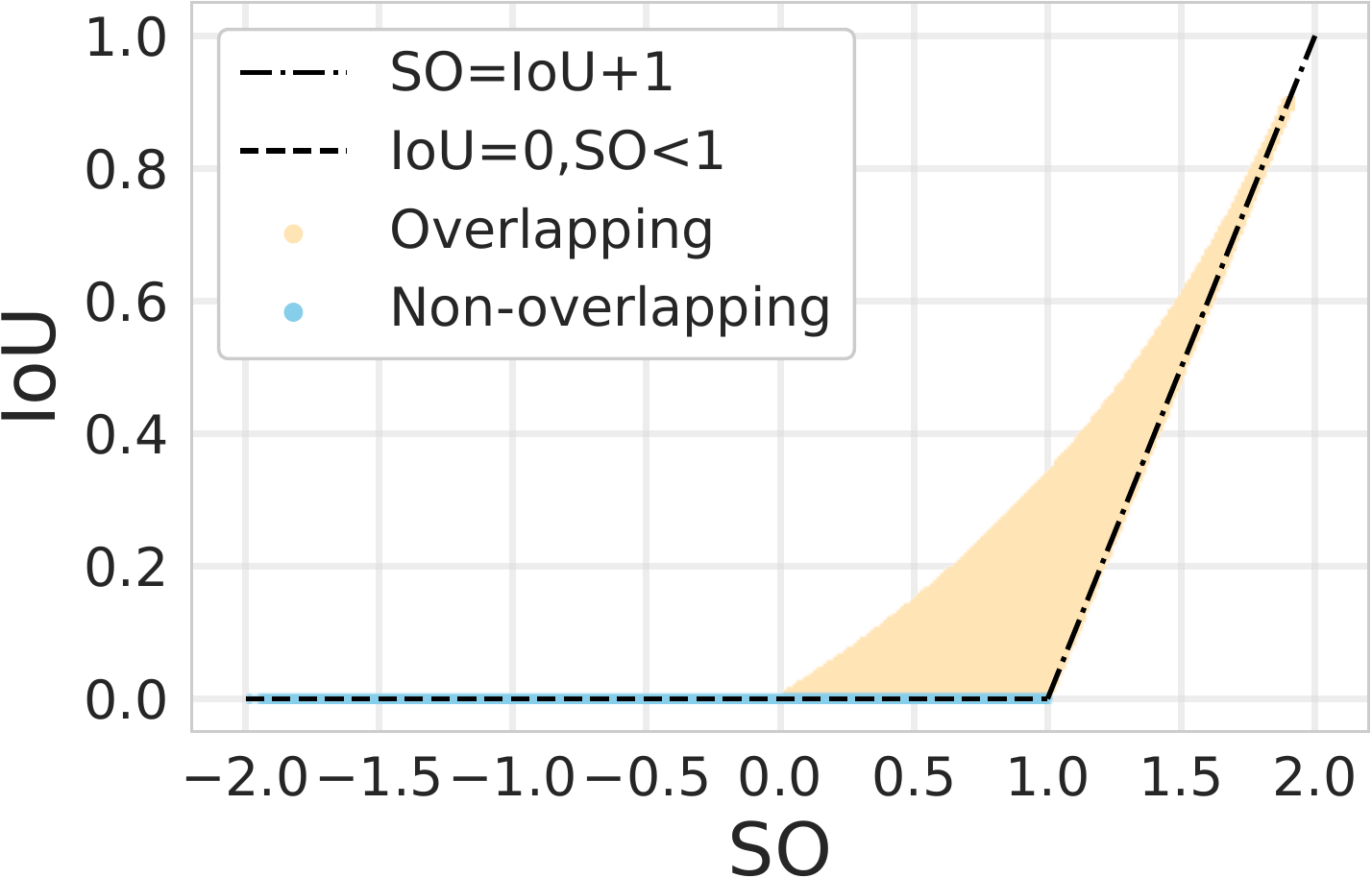}
\label{iou-so}
\end{subfigure}
\begin{subfigure}{0.49\columnwidth}
\includegraphics[width=0.99\linewidth]{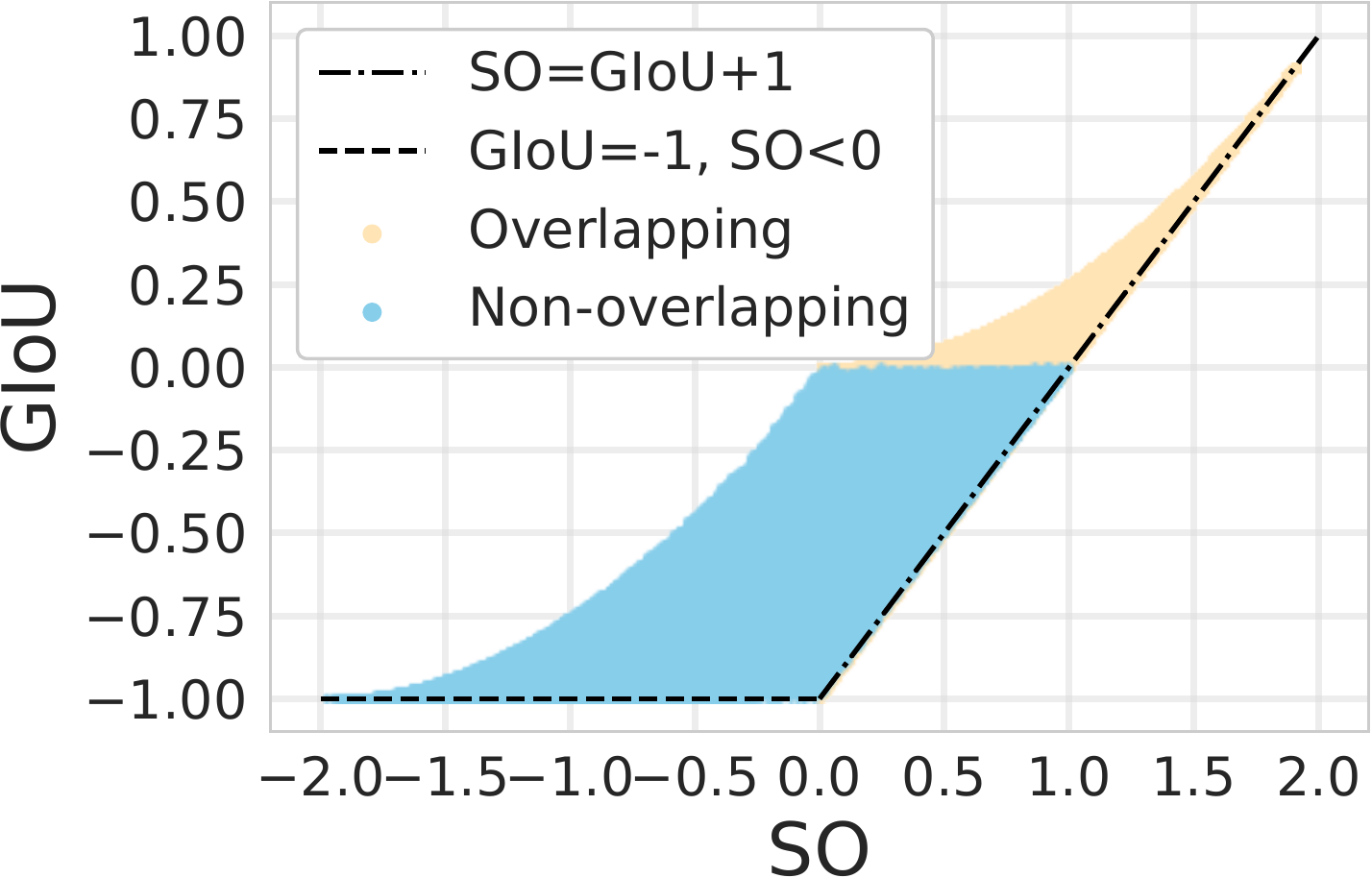}
\label{giou-so}
\end{subfigure}

\caption{
(left)~Relationship between IoU and SO, (right)~Relationship between GIoU
and SO for overlapping and non-overlapping samples.
}
\label{correlation}
\end{figure}

\subsection{Relationship with IoU and GIoU}\label{relationship}
For two arbitrary axis-aligned bounding boxes $A, B \in \mathcal{R}^4$, we can calculate the SO, IoU, and GIoU by their definitions respectively. The SO has the following properties: 
\begin{itemize}
    \item Similar to IoU and GIoU, SO is invariant to the scale of the problem. Because
the loss is normalized by the scale of box.
    \item
    SO is always a lower bound for $IoU + 1$ and $GIoU + 1$, and this lower bound becomes tighter when A and B have a stronger shape similarity.
    \item $\forall A, B \in \mathcal{R}^4 $, $0 \leq IoU(A, B) \leq 1$, SO and GIoU have a symmetric range, $\forall A, B  \in \mathcal{R}^4$, $-1 \leq GIoU(A, B) \leq 1$, $-2 \leq SO(A, B) \leq 2$.
    \begin{itemize}
        \item [1)] Similar to IoU and GIoU, the max value occurs when two objects match perfectly,
\emph{i.e.} if $ |A\cup B| = |A \cap B |$, then SO = 2, IoU = GIoU = 1.
        \item [2)] SO value asymptotically converges to -2 when two bounding boxes are far away. 
    \end{itemize}
    \item
    Different from IoU, SO still has gradient for non-overlaps cases and it has a larger gradient than GIoU.
\end{itemize}

We also demonstrate this correlation qualitatively in Fig. \ref{correlation} by taking over samples from 1000K random samples from coordinates of two 2D rectangles. It shows that SO has a strong correlation with IoU and GIoU in high IoU values. However, in the case of low overlapping, SO can make the bounding box change position and shape faster compared with IoU and GIoU. Thus, SO is promising to have a larger gradient in these cases. In conclusion, optimizing SO loss can be a better choice than optimizing IoU and GIoU loss.

\subsection{Simulation Experiment} \label{sim_exp}
\begin{figure}[!t]
\centering
\begin{subfigure}{0.23\textwidth}
\includegraphics[width=0.99\linewidth]{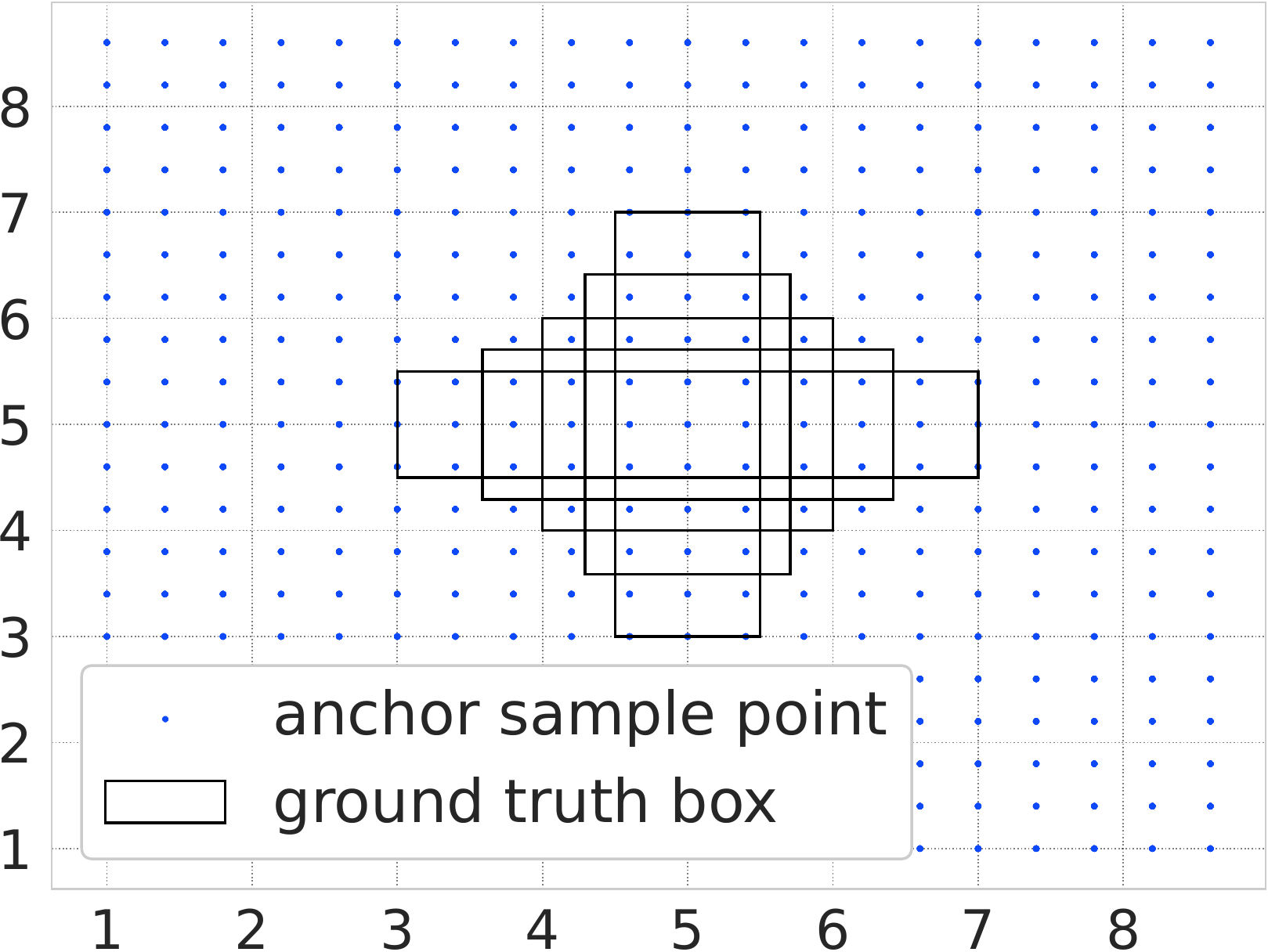}
\subcaption{anchors setting}
\label{anchors}
\end{subfigure}
\begin{subfigure}{0.23\textwidth}
\includegraphics[width=0.99\linewidth]{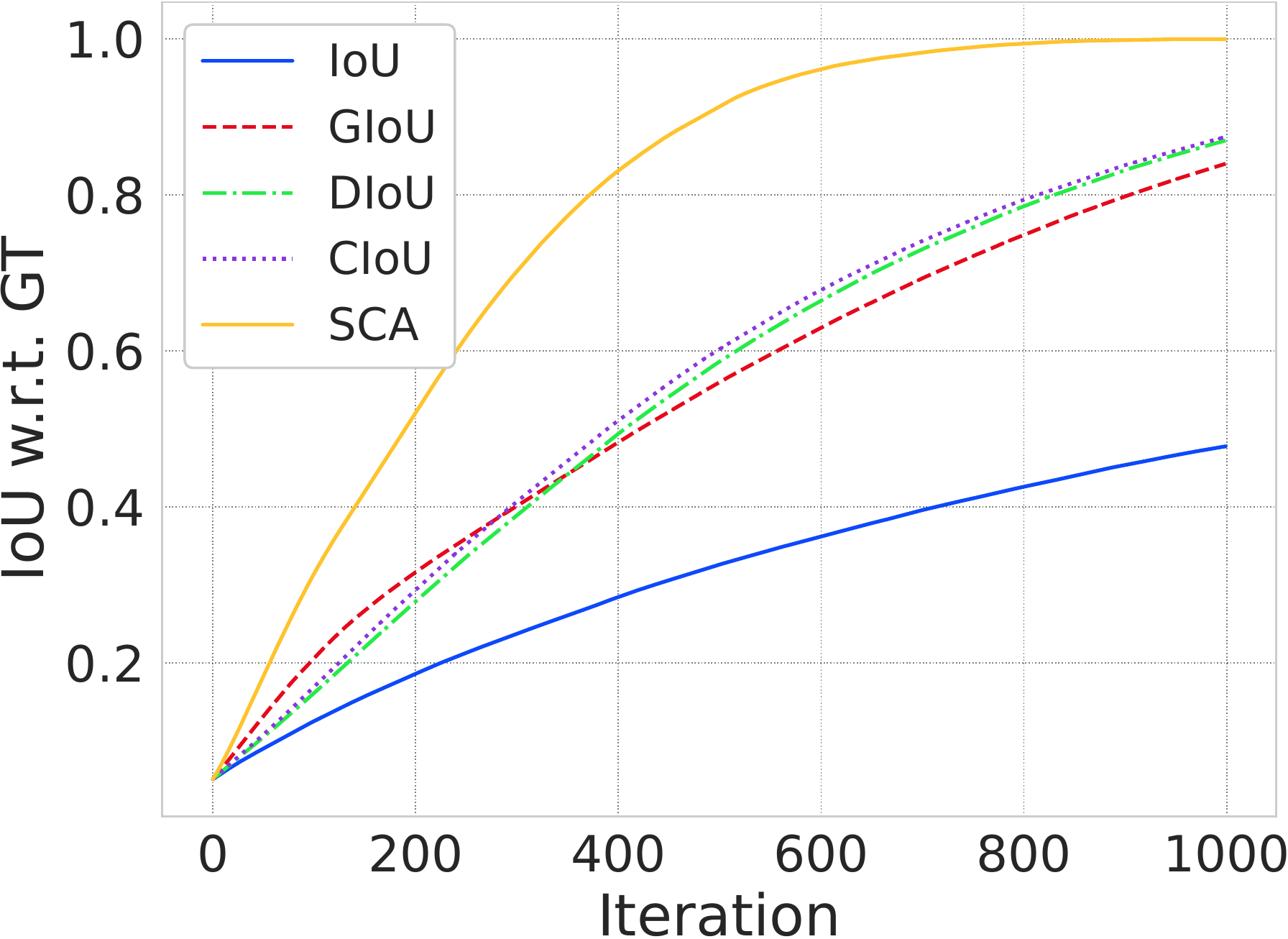}
\caption{regression results}
\label{reg_results}
\end{subfigure}

\caption{
(a) 3600 regression anchor boxes are adopted by
considering different scales and aspect ratios, (b) regression results curves with iteration t of different bounding box regression losses.
}
\label{toy_experiment_fig}
\end{figure}

To better understand the efficiency of our $\mathcal{L}_{SCA}$, we also provide a simple simulation experiment to compare $\mathcal{L}_{IoU}$, $\mathcal{L}_{GIoU}$, and $\mathcal{L}_{DIoU}$.
In the simulation experiment, we try to enumerate all possible anchor boxes. In particular, we choose 5 specific boxes with different aspect ratios (\emph{e.g.} 4:1, 2:1, 1:1, 1:2, 1:4) as ground truth boxes. Then anchor boxes are uniformly sampled in $20 \times 20$ grid with the ratio of (2:1, 1:1, 1:2) and scale of (2, 4, 6) and thus we have 3600 anchors as we can see in Fig.~\ref{anchors}. All the anchor boxes should be regressed to each ground truth box. 
Different with DIoU~\cite{zheng2019distance}, most boxes have overlap with their ground truth in our setting. Therefore, the simulation experiment is more similar to the real training procedure.

Given a loss function $\mathcal{L}$, we can simulate the procedure of bounding box regression using gradient descent algorithm. For each predicted box, we first calculate the loss function and the gradient:
\begin{align}
    Loss = \mathcal{L}(B^t, B_{gt}), \\
    \nabla B^t = \frac{\partial Loss}{\partial B^{t}},
\end{align}{}
Then, the update process can be obtained by:
\begin{align}
    B^t = B^{t-1} + \eta \nabla B^{t-1}.
\end{align}{}where $\eta$ is the learning rate, $B^t$ is the predicted box at iteration $t$ and $\nabla B^{t}$ is the gradient of the corresponding box. 
Finally, we calculate the mean IoU to evaluate the regression accuracy with different loss functions. The final regression result has shown in Fig. \ref{reg_results}.

SCALoss can converge faster from the results in Fig.~\ref{reg_results} because it produces large gradient for low overlapping boxes as we discuss in Sec.~\textit{Introduction}. This demonstrates SCALoss is a stricter similarity measure, and evaluating the similarity of corners and sides is more efficient in accurating bounding box regression.

\section{Experiment}

In this section, we construct experiments to evaluate the performance of our SCALoss by incorporating it into the most popular object detectors such as YOLOV3, SSD, Faster R-CNN. To this end, we replace their default regression losses with $\mathcal{L}_{SCA}$, \emph{i.e.}, we replace $\ell_n$ loss in YOLOV3 / SSD / Faster R-CNN.
We compare $\mathcal{L}_{SCA}$ against $\mathcal{L}_{IoU}$, $\mathcal{L}_{GIoU}$, $\mathcal{L}_{DIoU}$, and $\mathcal{L}_{CIoU}$. We use the mmdetection~\cite{chen2019mmdetection} toolbox to conduct all our experiments except YOLOV3. We use Pytorch~\cite{paszke2019pytorch} with the NVIDIA 1080Ti GPU in Ubuntu. All models are pre-trained on ImageNet~\cite{deng2009imagenet}.


\subsection{Dataset} 
All results are reported on three popular object detection benchmarks, the PASCAL
VOC, MS-COCO, and LVIS. \\
\textit{PASCAL VOC:} The Pascal Visual Object Classes (VOC)  dataset is one of the most popular benchmarks for category classification, detection, and semantic segmentation. For object detection, it has 20 pre-defined classes with annotated bounding boxes. We use PASCAL VOC 2007 + 2012 (the union of VOC 2007 and VOC 2012 trainval) with 16551 images as the training set and PASCAL VOC 2007 test with 4952 images as the test set.\\
\textit{MS COCO:} 
Microsoft Common Objects in Context (MS-COCO) is another popular dataset for object detection, instance segmentation, and object keypoint detection. It is a large scale dataset with 80 pre-defined classes. We use COCO $\textit{train2017}$ with 135k images as the training set, $\textit{val2017}$ with 5k images as the validation set and  $\textit{test-dev}$ with 20k images as the test set.\\
\textit{LVIS:} LVIS is a large vocabulary dataset for instance segmentation, which contains 1203 categories in current version v1.0.
In LVIS, categories are divided into three groups according to the number of images that contains those categories: rare
(1-10 images), common (11-100), and frequent ($>$100). We train our model on 57k train images and evaluate it on 20k
val set. 

\subsection{Evaluation}
In this paper, we adopt the same mAP calculation method as MS COCO to report all our results. The mAP score is calculated by taking mean AP over all classes and over all 10 IoU thresholds, \emph{i.e.} IoU= {0.5, 0.55, ..., 0.95}. While PASCAL VOC only considers one IoU, \emph{i.e.}, IoU = 0.5, we modify it same as COCO for better performance comparison. For LVIS, we also report the $AP_r, AP_c, AP_f$ for rate, common, and frequent categories respectively. 

\subsection{YOLOV3 and SSD}
We first use two representative one-stage detectors, \emph{i.e.}, YOLOV3 and SSD, to construct experiments. 

\subsubsection{YOLOV3-tiny on COCO}
Following its training protocol\footnote{\url{https://github.com/ultralytics/yolov3}}, we train YOLOV3-tiny with every aforementioned bounding box regression loss on the training set for 300 epochs.  The input image size is $640 \times 640$. We show the performance of each loss in Table \ref{yolo-tiny_coco}. The result shows that training YOLOV3-tiny with $\mathcal{L}_{SCA}$ can considerably improve its performance, near 1 AP, compared to $\mathcal{L}_{IoU}$, $\mathcal{L}_{GIoU}$, $\mathcal{L}_{DIoU}$, and $\mathcal{L}_{CIoU}$. Note that the improvement mostly comes from high overlap metrics, \emph{e.g.}, near 1.8 points in AP75. Our method promotes the gradient for low overlapping cases, so the APs from low overlap metrics~(like AP50, AP65) are much better than $\mathcal{L}_{IoU}$, $\mathcal{L}_{GIoU}$, $\mathcal{L}_{DIoU}$, and $\mathcal{L}_{CIoU}$.

\begin{figure}[!t]
\centering
\includegraphics[width=0.46\textwidth]{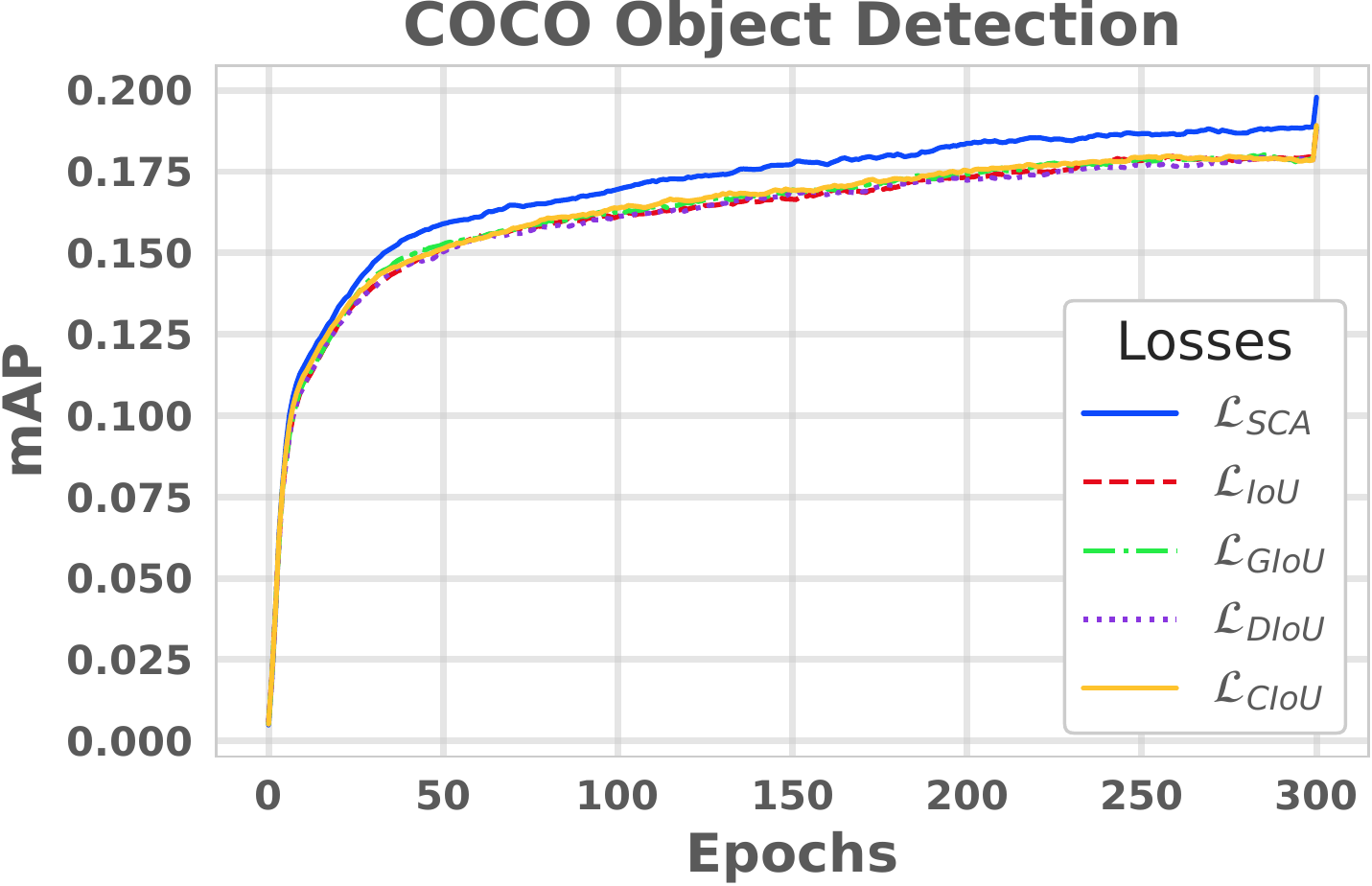}
 \caption{mAP value against train epochs for YOLO-tiny using $\mathcal{L}_{GIoU}$, $\mathcal{L}_{DIoU}$, and $\mathcal{L}_{SCA}$ on COCO 2017 \textit{val} set. }
 \label{yolo-tiny_epoch-map}
\end{figure}

We also plot the relationship between training epoch and mAP, as we can see in Fig.~\ref{yolo-tiny_epoch-map}, our SCALoss can converge faster than other losses and get consistent higher AP during the training process.

\subsubsection{YOLOV3 on COCO}
Similarly, we train YOLOV3 using each of aforementioned bounding box regression losses on COCO dataset. The backbone is darknet-53 and other settings are same as YOLOv3-tiny. As Table~\ref{yolo_coco} shows, SCALoss can also surpass other losses consistently for the higher baseline models.

\begin{table}[h!]

\centering
\resizebox{.98\columnwidth}{!}{
\begin{spreadtab}{{tabular}{c c c c c c c}}
 \toprule
 @Loss & @mAP & @AP$_{50}$ & @ AP$_{65}$ & @ AP$_{75}$ &@ AP$_{80}$ & @AP$_{90}$ \\
 \midrule\midrule
 $@ \mathcal{L}_{IoU}$ & 18.8 & 36.2 & 27.2 & 17.3 & 11.6 & 1.9\\
 \midrule
 $@ \mathcal{L}_{GIoU}$ & 18.8 & 36.2 & 27.1 & 17.6 & 11.8 & 2.1 \\
 @rela. improv. & :={round((b3-b2)/b2*100,2)}\% & :={round((c3-c2)/c2*100,2)}\% & :={round((d3-d2)/d2*100,2)}\% & :={round((e3-e2)/e2*100,2)}\% & :={round((f3-f2)/f2*100,2)}\% & :={round((g3-g2)/g2*100,2)}\% \\
 \midrule
 $@ \mathcal{L}_{DIoU}$ & 18.8 & 36.4 & 26.9 & 17.2 &  11.8 & 1.9\\
 @rela. improv.  & :={round((b5-b2)/b2*100,2)}\% & :={round((c5-c2)/c2*100,2)}\% & :={round((d5-d2)/d2*100,2)}\% & :={round((e5-e2)/e2*100,2)}\% & :={round((f5-f2)/f2*100,2)}\% & :={round((g5-g2)/g2*100,2)}\% \\
 \midrule
 $@ \mathcal{L}_{CIoU}$ & 18.9 & 36.6 & 27.3 & 17.2 & 11.6 & 2.1 \\
 @rela. improv.  & :={round((b7-b2)/b2*100,2)}\% & :={round((c7-c2)/c2*100,2)}\% & :={round((d7-d2)/d2*100,2)}\% & :={round((e7-e2)/e2*100,2)}\% & :={round((f7-f2)/f2*100,2)}\% & :={round((g7-g2)/g2*100,2)}\% \\
 \midrule
 $@ \mathcal{L}_{SCA}$ & {\fontseries{b}\selectfont}\STcopy{>}{19.9} & {\fontseries{b}\selectfont}\STcopy{>}{36.6} & {\fontseries{b}\selectfont}\STcopy{>}{28.3} &
 {\fontseries{b}\selectfont}\STcopy{>}{19.1} & {\fontseries{b}\selectfont}\STcopy{>}{13.3} & {\fontseries{b}\selectfont}\STcopy{>}{2.7} \\
 @rela. improv. & \textbf{:={round((b9-b2)/b2*100,2)}}\% & \textbf{:={round((c9-c2)/c2*100,2)}}\% &
 \textbf{:={round((d9-d2)/d2*100,2)}}\% & \textbf{:={round((e9-e2)/e2*100,2)}}\% & \textbf{:={round((f9-f2)/f2*100,2)}}\% & \textbf{:={round((g9-g2)/g2*100,2)}}\%\\
 \bottomrule
 \end{spreadtab}
 }
  \caption{Comparison between the performance of YOLOV3-tiny trained using
 $\mathcal{L}_{IoU}$, $\mathcal{L}_{GIoU}$, $\mathcal{L}_{DIoU}$, $\mathcal{L}_{CIoU}$, and $\mathcal{L}_{SCA}$ losses on COCO 2017 \textit{val} set.}
\label{yolo-tiny_coco}
\end{table}

\begin{table}[h!]

\centering
\resizebox{.98\columnwidth}{!}{
\begin{spreadtab}{{tabular}{c c c c c c c}}
 \toprule
 @Loss & @mAP & @AP$_{50}$ & @ AP$_{65}$ & @ AP$_{75}$ &@ AP$_{80}$ & @AP$_{90}$ \\
 \midrule\midrule
 $@ \mathcal{L}_{IoU}$ & 44.8 & 64.2 & 57.5 & 48.8 & 41.8 & 20.7 \\
 \midrule
 $@ \mathcal{L}_{GIoU}$ & 44.7 & {\fontseries{b}\selectfont}\STcopy{>}{64.4} & 57.5 & 48.5 & 42.0 & 20.4 \\
 @rela. improv.  & :={round((b3-b2)/b2*100,2)}\% & :={round((c3-c2)/c2*100,2)}\% & :={round((d3-d2)/d2*100,2)}\% & :={round((e3-e2)/e2*100,2)}\% & :={round((f3-f2)/f2*100,2)}\% & :={round((g3-g2)/g2*100,2)}\% \\
 \midrule
 $@ \mathcal{L}_{DIoU}$ & 44.7 & 64.3 & 57.5 & 48.9 & 42.1 & 19.8 \\
 @rela. improv.  & :={round((b5-b2)/b2*100,2)}\% & :={round((c5-c2)/c2*100,2)}\% & :={round((d5-d2)/d2*100,2)}\% & :={round((e5-e2)/e2*100,2)}\% & :={round((f5-f2)/f2*100,2)}\% & :={round((g5-g2)/g2*100,2)}\% \\
 \midrule
 $@ \mathcal{L}_{CIoU}$ & 44.7 & 64.3& 57.5 & 48.9 & 41.7 & 19.8 \\
 @rela. improv.  & :={round((b7-b2)/b2*100,2)}\% & :={round((c7-c2)/c2*100,2)}\% & :={round((d7-d2)/d2*100,2)}\% & :={round((e7-e2)/e2*100,2)}\% & :={round((f7-f2)/f2*100,2)}\% & :={round((g7-g2)/g2*100,2)}\% \\
 \midrule
 $@ \mathcal{L}_{SCA}$ & {\fontseries{b}\selectfont}\STcopy{>}{45.3} & 64.1 & {\fontseries{b}\selectfont}\STcopy{>}{57.9} &
 {\fontseries{b}\selectfont}\STcopy{>}{49.9} & {\fontseries{b}\selectfont}\STcopy{>}{43.3} & {\fontseries{b}\selectfont}\STcopy{>}{21.4} \\
 @rela. improv. & \textbf{:={round((b9-b2)/b2*100,2)}}\% & :={round((c9-c2)/c2*100,2)}\% &
 \textbf{:={round((d9-d2)/d2*100,2)}}\% & \textbf{:={round((e9-e2)/e2*100,2)}}\% & \textbf{:={round((f9-f2)/f2*100,2)}}\% & \textbf{:={round((g9-g2)/g2*100,2)}}\%\\
 \bottomrule
 \end{spreadtab}
 }
  \caption{Comparison between the performance of YOLOV3 trained using
 $\mathcal{L}_{IoU}$, $\mathcal{L}_{GIoU}$, $\mathcal{L}_{DIoU}$, $\mathcal{L}_{CIoU}$, and $\mathcal{L}_{SCA}$ losses on COCO 2017 \textit{val} set.}
\label{yolo_coco}
\end{table}

 \begin{figure*}[t!]
  \centering
  
   \includegraphics[width=.20\linewidth, height=2.20cm]{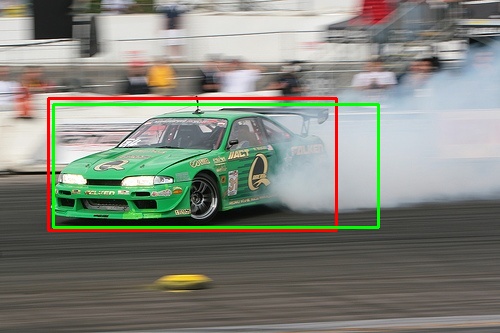}
  \hspace{0.01\textwidth}
   \includegraphics[width=.20\linewidth, height=2.20cm]{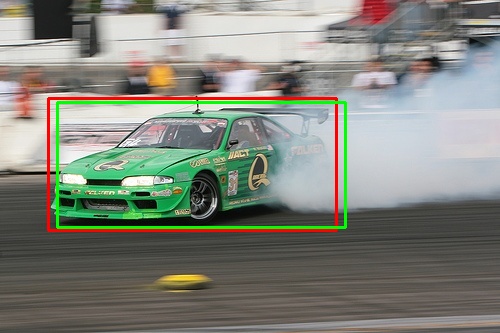}
  \hspace{0.01\textwidth}
  \includegraphics[width=.20\textwidth, height=2.20cm]{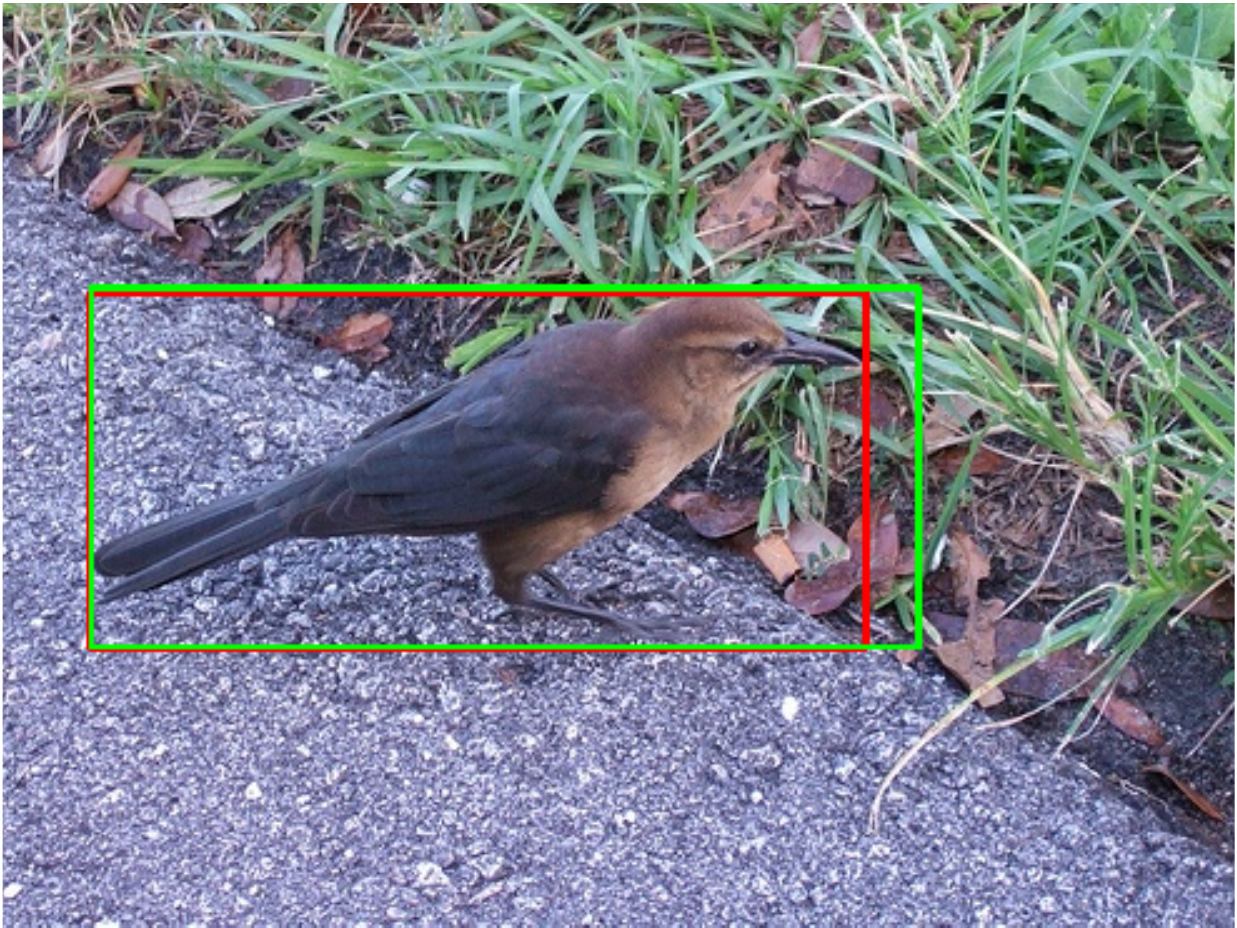}
  \hspace{0.01\textwidth}
  \includegraphics[width=.20\textwidth, height=2.20cm]{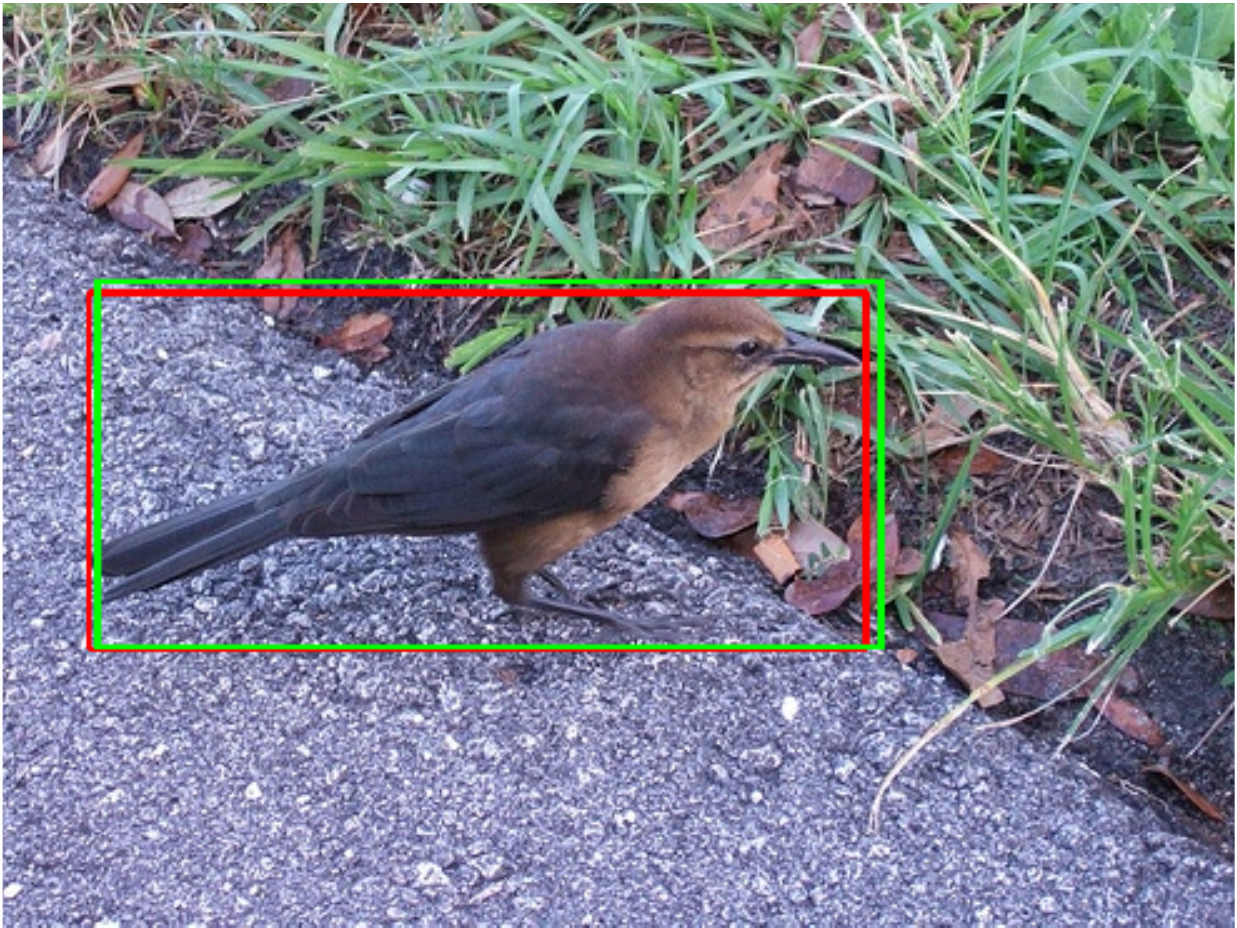}
  
  \vspace{5pt}

  \includegraphics[width=.20\textwidth, height=2.20cm]{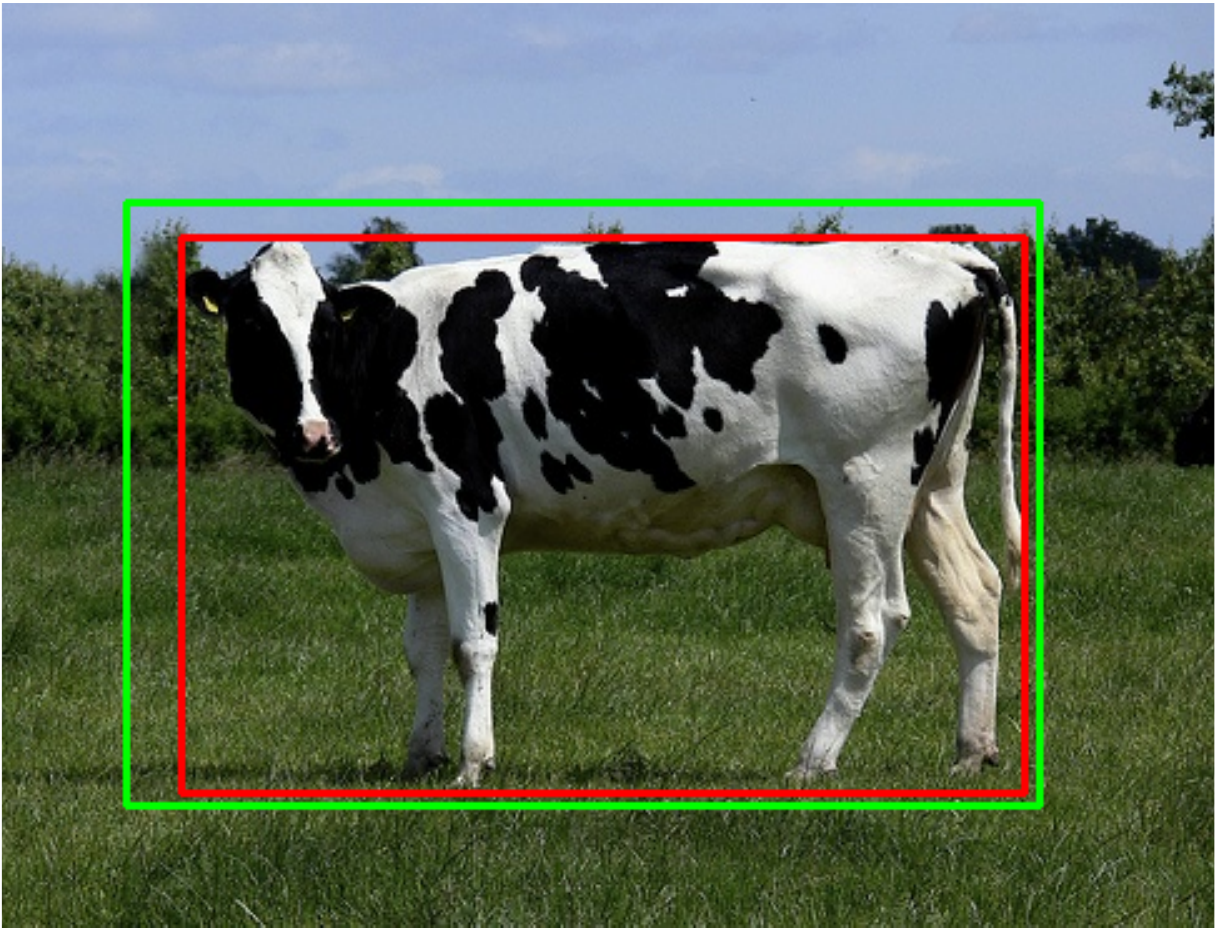}
  \hspace{0.01\textwidth}
  \includegraphics[width=.20\textwidth, height=2.20cm]{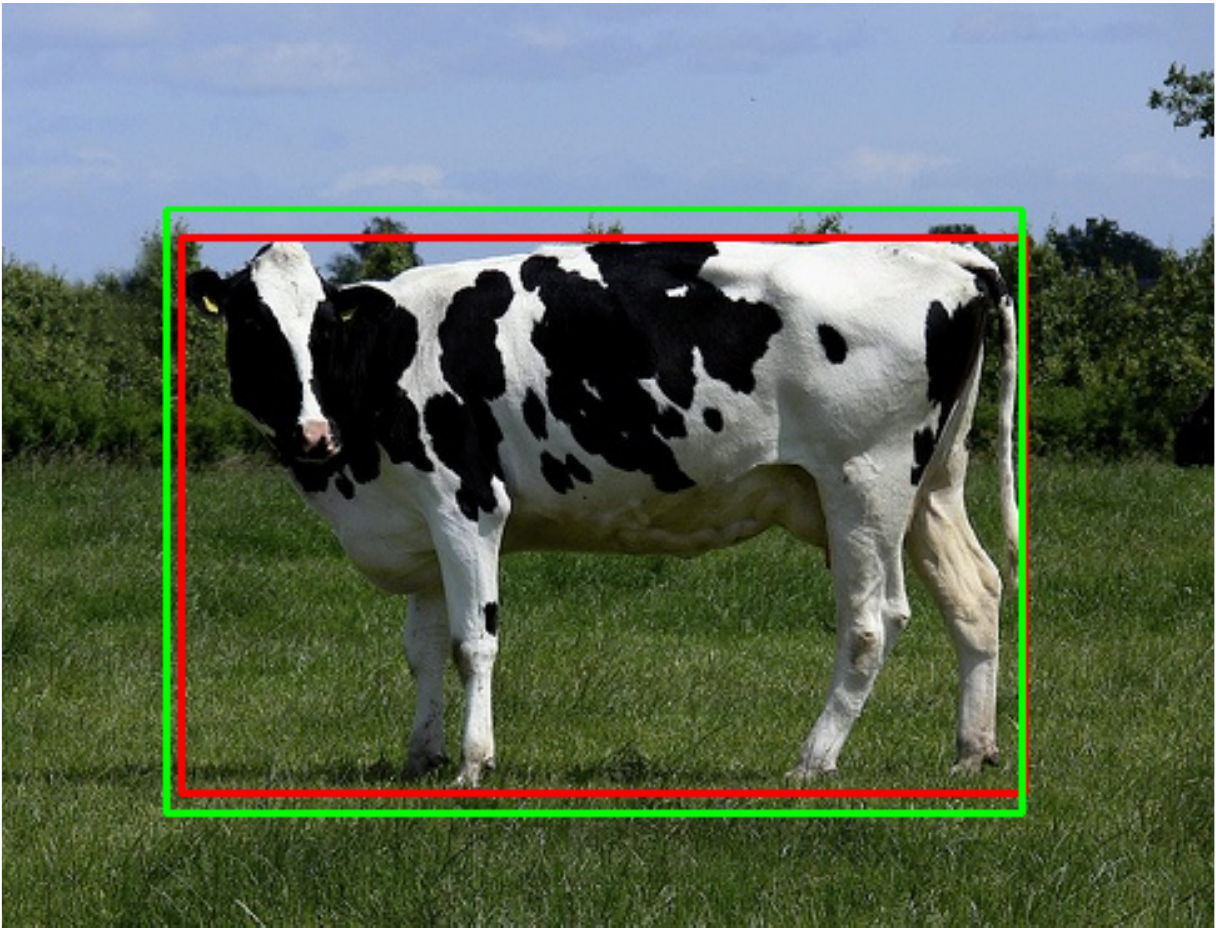}
  \hspace{0.01\textwidth}
  \includegraphics[width=.20\textwidth, height=2.20cm]{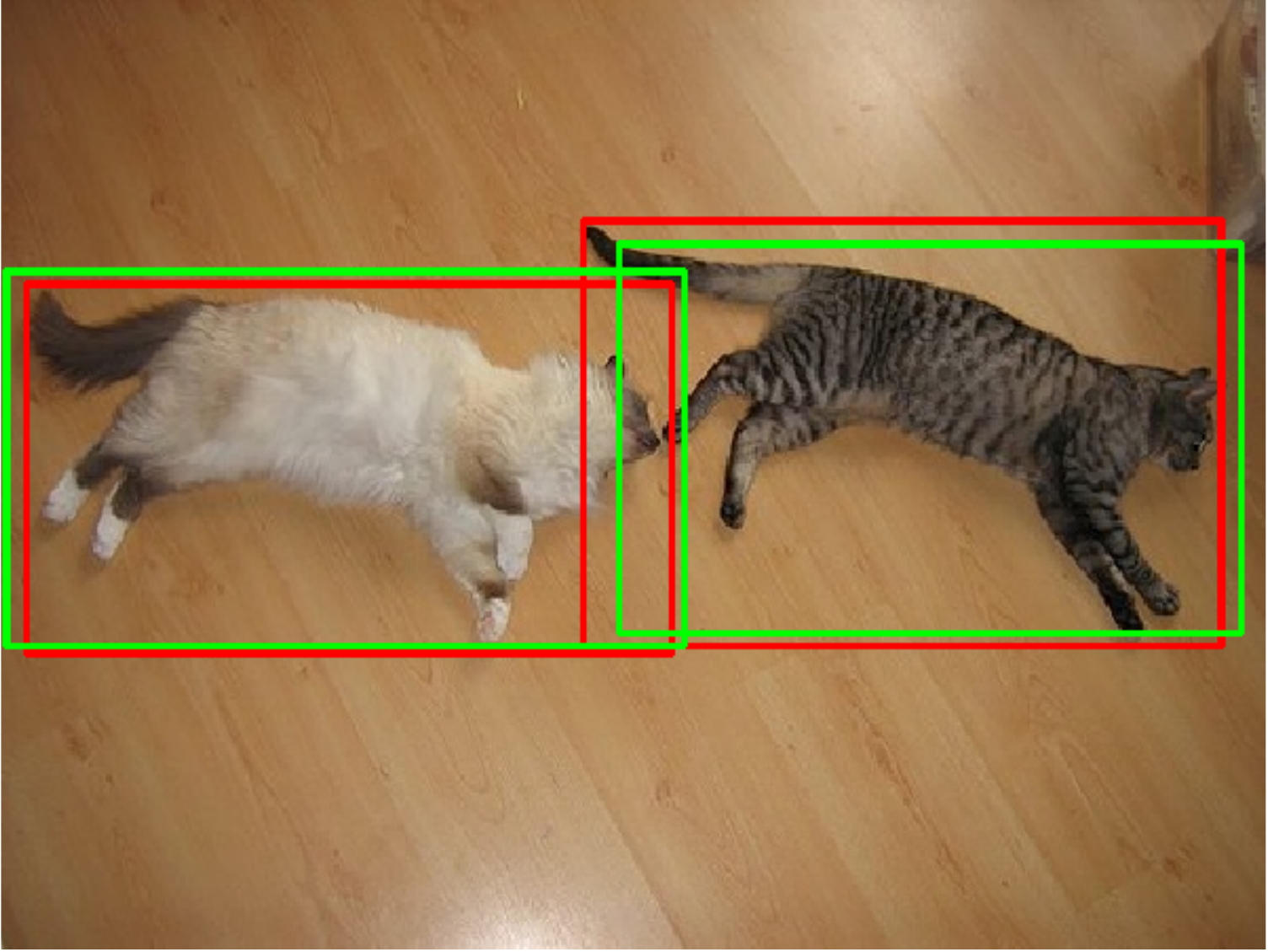}
  \hspace{0.01\textwidth}
  \includegraphics[width=.20\textwidth, height=2.20cm]{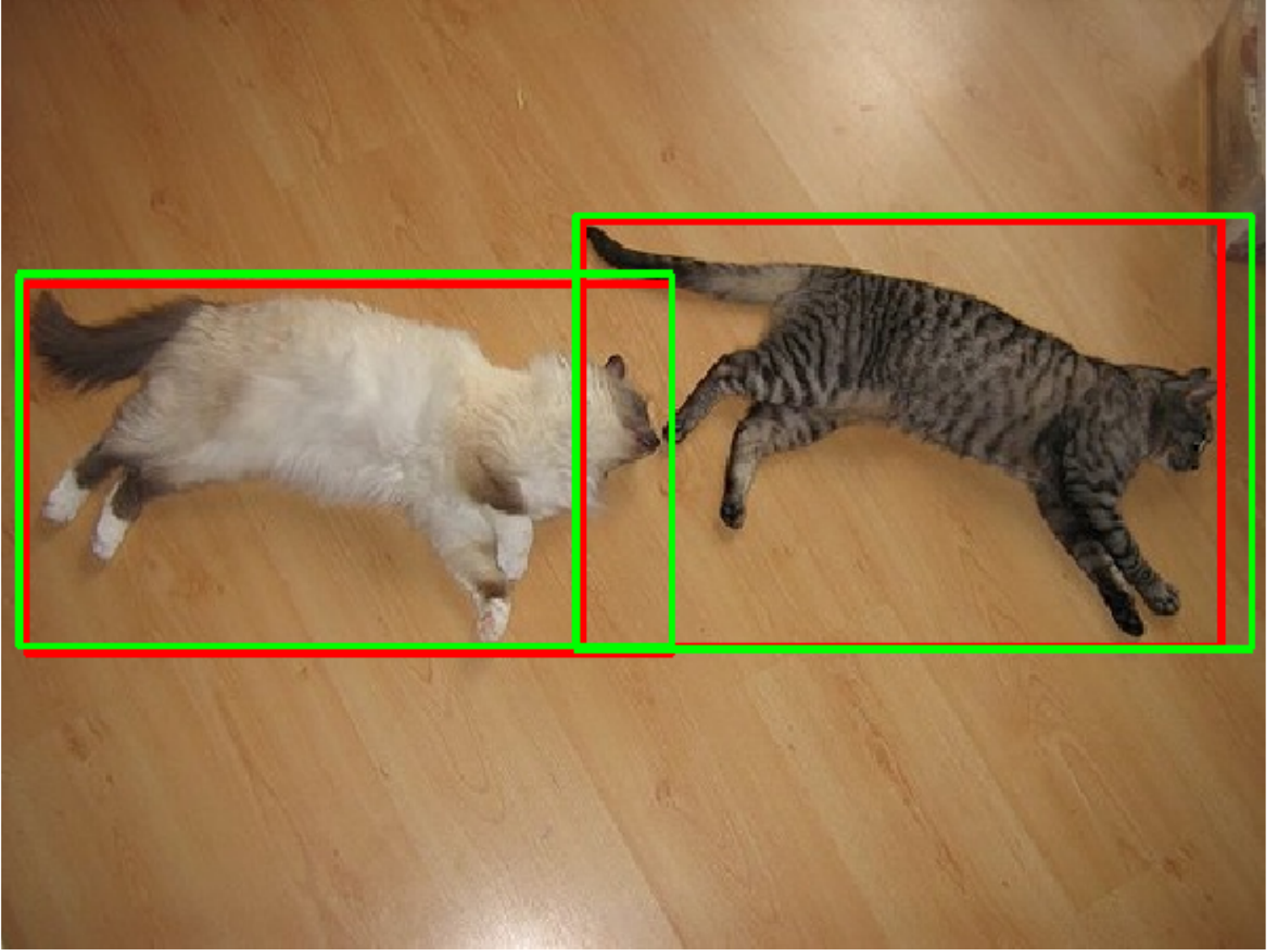}

  \vspace{5pt}

  \includegraphics[width=.20\textwidth]{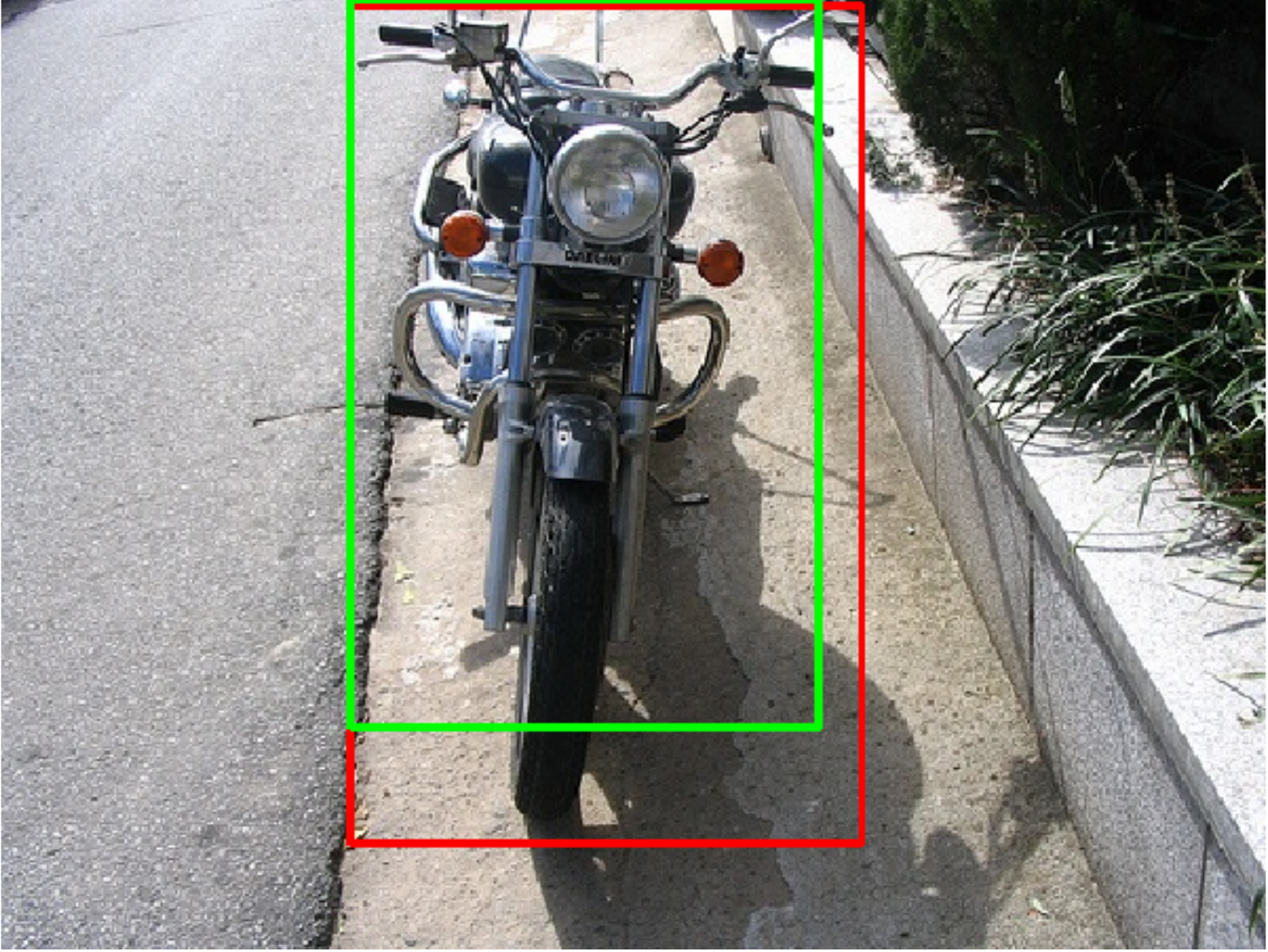}
  \hspace{0.01\textwidth}
  \includegraphics[width=.20\textwidth]{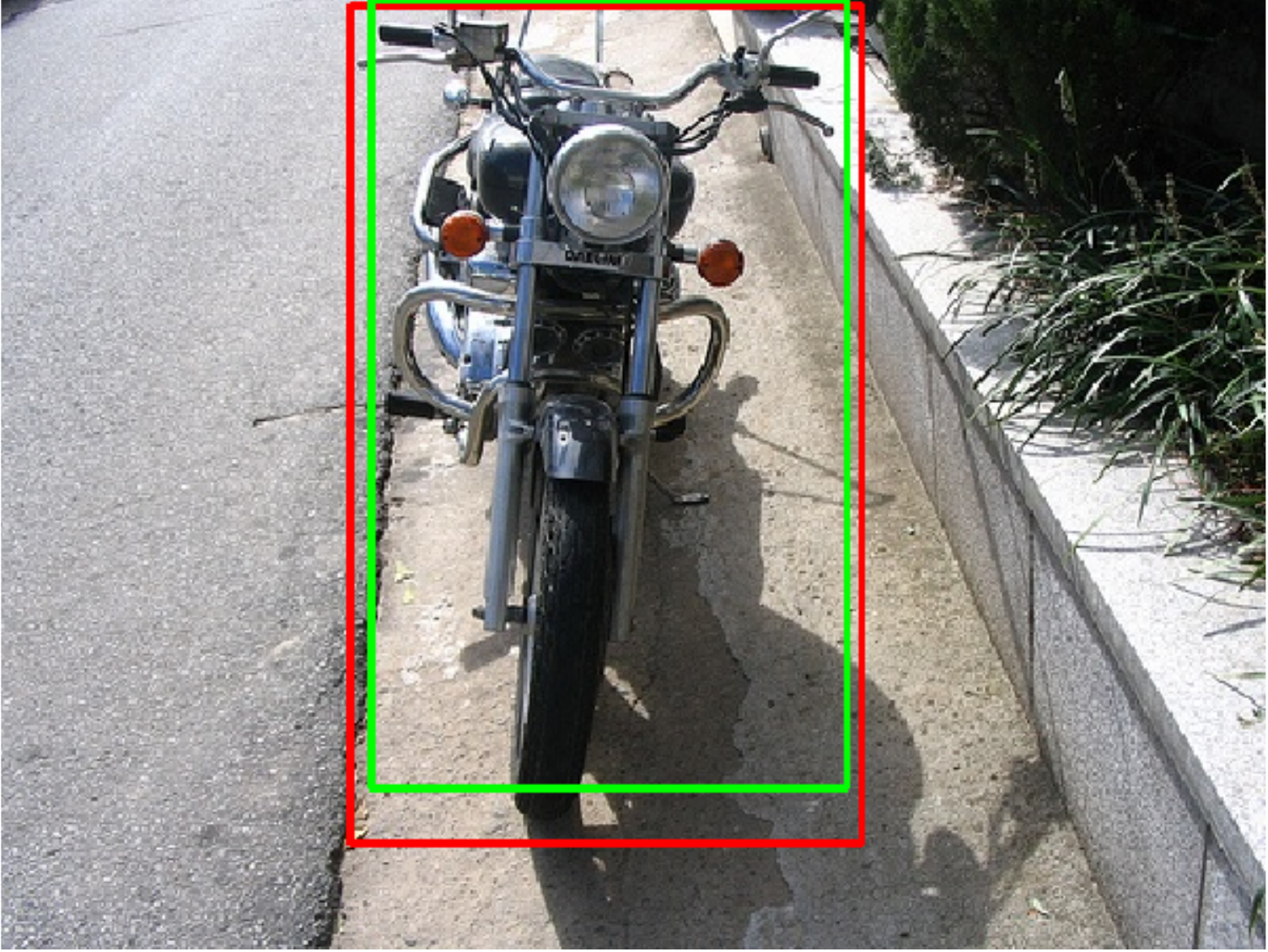}
  \hspace{0.01\textwidth}
  \includegraphics[width=.20\textwidth]{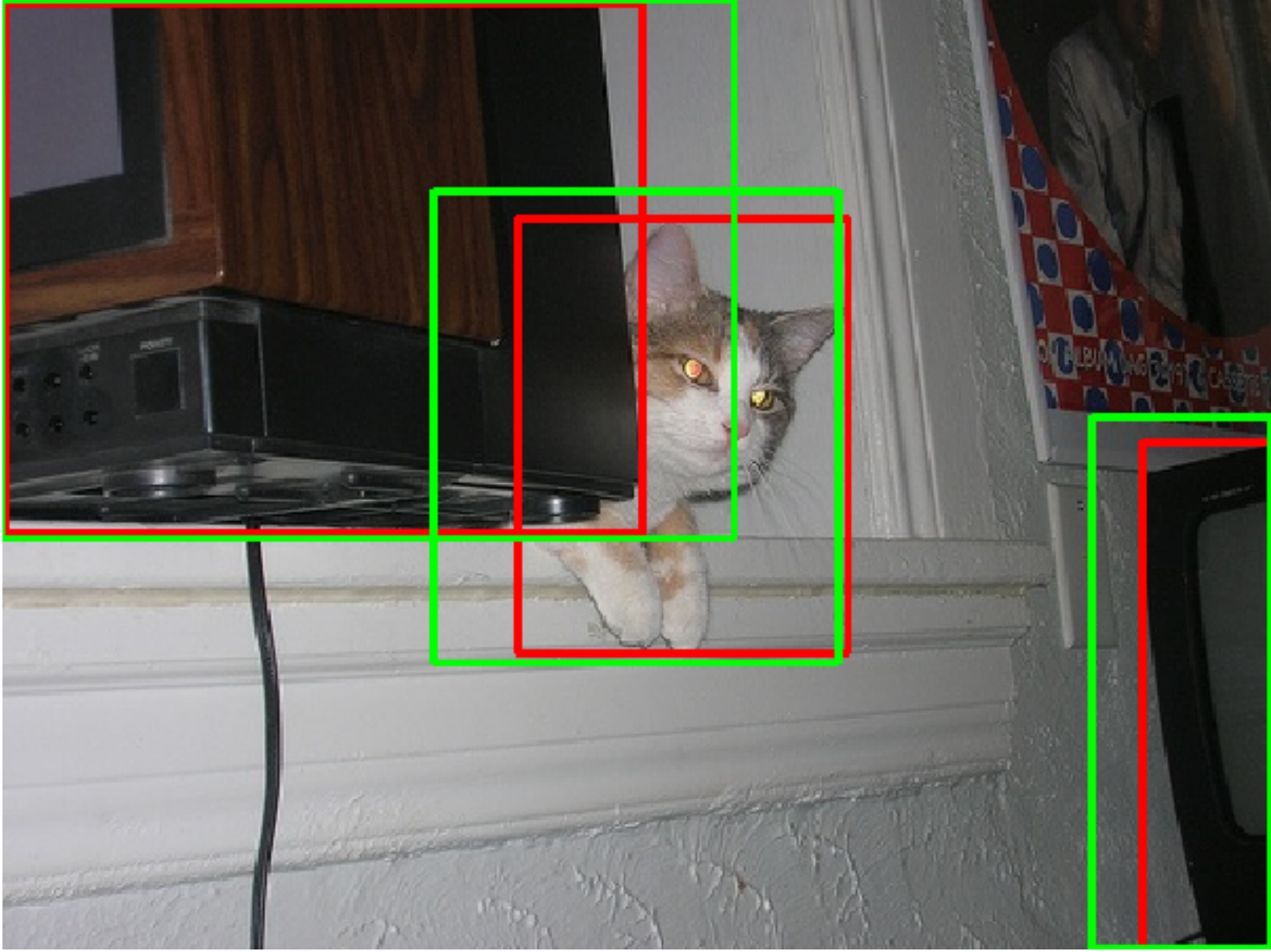}
  \hspace{0.01\textwidth}
  \includegraphics[width=.20\textwidth]{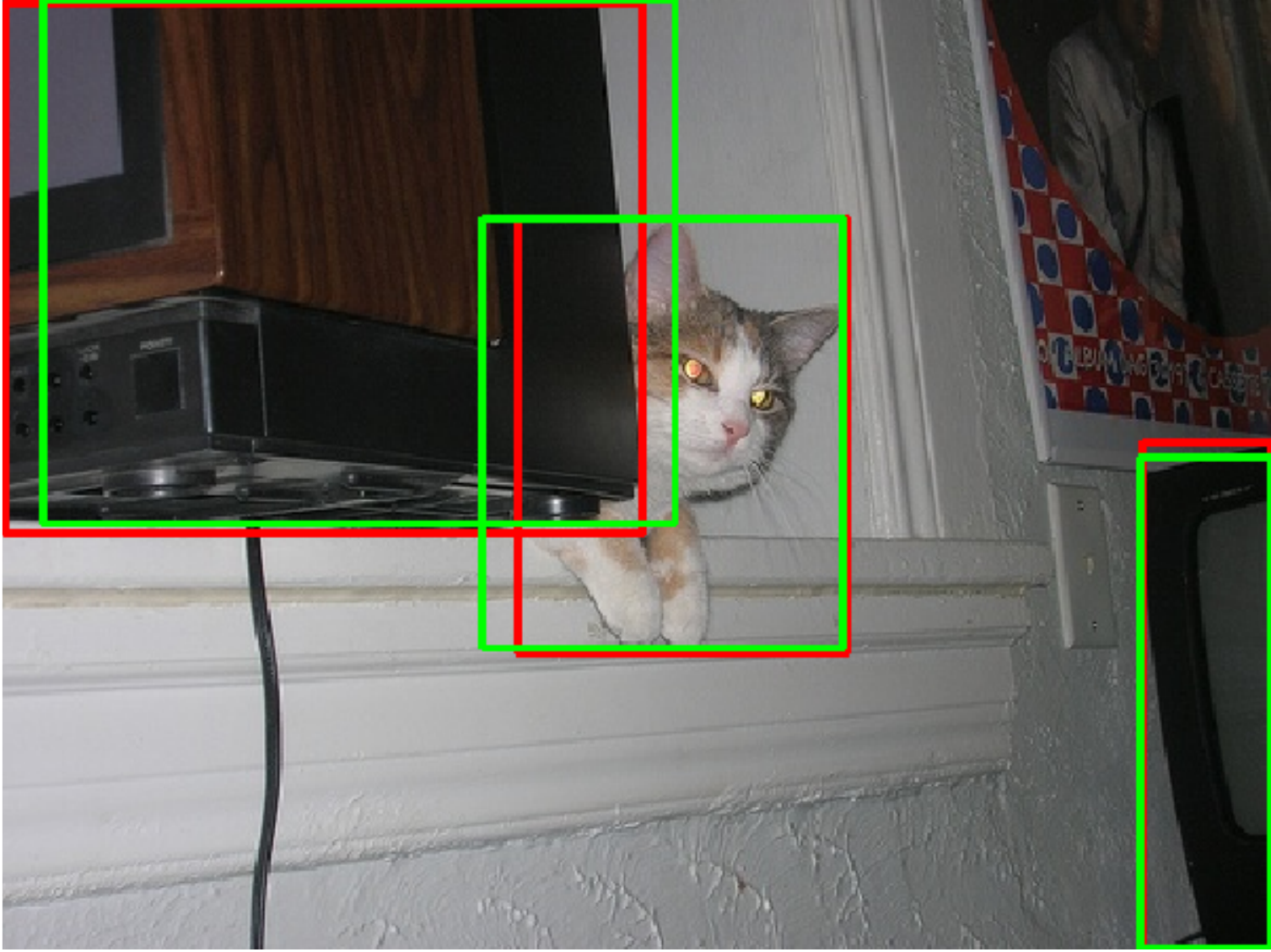}
  \caption{Examples results from PASCAL VOC dataset using Faster R-CNN trained with $\mathcal{L}_{GIoU}$ and $\mathcal{L}_{SCA} $~(left to right). Red: the ground truth box, green: the predicted box.}\label{detection_result}
\end{figure*}

\subsubsection{SSD on PASCAL VOC}
When training SSD on PASCAL VOC, we use the same setting as the COCO. The training epochs is 72.
The performance for each loss has been shown in Table \ref{ssd_voc}. The result shows that training SSD with $\mathcal{L}_{SCA}$ can considerably improve its performance compared to $\mathcal{L}_{IoU}$. Moreover, $\mathcal{L}_{SCA}$ can get better performance than $\mathcal{L}_{GIoU}$, and $\mathcal{L}_{DIoU}$.

\begin{table}[h!]

\centering

\resizebox{.98\columnwidth}{!}{
\begin{spreadtab}{{tabular}{c c c c c c c}}
 \toprule
 @Loss & @mAP & @AP$_{50}$ & @ AP$_{65}$ & @ AP$_{75}$ &@ AP$_{80}$ & @AP$_{90}$ \\
 \midrule\midrule
 @$\mathcal{L}_{IoU}$ & 52.28 & 78.26 & 68.51 & 56.22 & 46.90 & 20.20 \\
 \midrule\midrule
 @$\mathcal{L}_{GIoU}$ & 52.50 & 78.57 & 69.11 & 56.71 & 46.87 & 19.88\\
 @rela. improv.  & :={round((b3-b2)/b2*100,2)}\% & :={round((c3-c2)/c2*100,2)}\%
 & :={round((d3-d2)/d2*100,2)}\% & :={round((e3-e2)/e2*100,2)}\% & :={round((f3-f2)/f2*100,2)}\% & :={round((g3-g2)/g2*100,2)}\% \\
 \midrule
 @$\mathcal{L}_{DIoU}$ & 52.65 & 78.64  & 69.11 & 56.55 & 47.9 & 20.09 \\
 @rela. improv.  & :={round((b5-b2)/b2*100,2)}\% & :={round((c5-c2)/c2*100,2)}\%
 & :={round((d5-d2)/d2*100,2)}\% & :={round((e5-e2)/e2*100,2)}\% & :={round((f5-f2)/f2*100,2)}\% & :={round((g5-g2)/g2*100,2)}\% \\
 \midrule
 @$\mathcal{L}_{CIoU}$ & 52.75 & 78.76 & 69.03 & 56.45 & 48.56 & 20.69 \\
 @rela. improv.  & :={round((b7-b2)/b2*100,2)}\% & :={round((c7-c2)/c2*100,2)}\%
 & :={round((d7-d2)/d2*100,2)}\% & :={round((e7-e2)/e2*100,2)}\% & :={round((f7-f2)/f2*100,2)}\% & :={round((g7-g2)/g2*100,2)}\% \\
 \midrule
 @$\mathcal{L}_{SCA}$ &  {\fontseries{b}\selectfont}\STcopy{>}{53.45}  & {\fontseries{b}\selectfont}\STcopy{>}{78.82} & {\fontseries{b}\selectfont}\STcopy{>}{69.17} & {\fontseries{b}\selectfont}\STcopy{>}{57.22} & {\fontseries{b}\selectfont}\STcopy{>}{48.49} & {\fontseries{b}\selectfont}\STcopy{>}{22.67} \\
 @rela. improv. & \textbf{:={round((b9-b2)/b2*100,2)}}\% & \textbf{:={round((c9-c2)/c2*100,2)}}\% &
 \textbf{:={round((d9-d2)/d2*100,2)}}\% & \textbf{:={round((e9-e2)/e2*100,2)}}\% & \textbf{:={round((f9-f2)/f2*100,2)}}\% & \textbf{:={round((g9-g2)/g2*100,2)}}\%\\
 \bottomrule
\end{spreadtab}}
 \caption{Comparison between the performance of SSD trained using $\mathcal{L}_{IoU}$, $\mathcal{L}_{GIoU}$, $\mathcal{L}_{DIoU}$, $\mathcal{L}_{CIoU}$, and $\mathcal{L}_{SCA}$ losses on the PASCAL VOC 2007 \textit{test} set.
 }
 \label{ssd_voc}
\end{table}

\subsection{Faster R-CNN}
Faster R-CNN is a two-stage detector, which generates object proposals for the second stage to classify and refine bounding boxes. We use the ResNet50-FPN backbone network and replace the $\ell_1$-smooth loss in the second stage in Faster R-CNN. 

\paragraph{Faster R-CNN on PASCAL VOC} We train Faster R-CNN for 12 epochs on PASCAL VOC dataset, and the input image is resized to $1000 \times 600$. The final results have been reported in Table \ref{faster_voc}. The results show that training Faster-RCNN using our SCALoss can consistently improve its performance compared to IoU loss (near 2\%).  SCALoss can improve the performance with gains of near 0.9 AP / 0.8 AP than GIoU loss, CIoU loss respectively.

Fig.~\ref{detection_result} shows the qualitative results of models trained using GIoU loss and SCA loss. Adopting SCA loss can get more accurate bounding boxes than GIoU loss and the corners of bounding boxes can be regressed better, which demonstrates SCA can better align the bounding box and yield a better detection performance.

\begin{table}[h!]

 \centering
\resizebox{.98\columnwidth}{!}{
\begin{spreadtab}{{tabular}{c c c c c c c}}
 \toprule
 @Loss & @mAP & @AP$_{50}$ & @ AP$_{65}$ & @ AP$_{75}$ &@ AP$_{80}$ & @AP$_{90}$ \\
 \midrule\midrule
  @ $\mathcal{L}_{IoU}$ & 50.85 & 79.60 & 69.85 & 55.14 & 43.23 & 13.01 \\
  \midrule
  @ $\mathcal{L}_{GIoU}$ & 50.90 & 79.69 & 70.56 & 55.00 & 43.34 & 12.70 \\
  @ rela. improv. & :={round((b3-b2)/b2*100,2)}\% & :={round((c3-c2)/c2*100,2)}\%
  & :={round((d3-d2)/d2*100,2)}\% & :={round((e3-e2)/e2*100,2)}\% & :={round((f3-f2)/f2*100,2)}\% & :={round((g3-g2)/g2*100,2)}\% \\
  \midrule
  @ $\mathcal{L}_{DIoU}$ & 50.86 & 79.99 & 70.48 & 54.56 & 42.79 & 12.80 \\
  @ rela. improv.  & :={round((b5-b2)/b2*100,2)}\% & :={round((c5-c2)/c2*100,2)}\%
  & :={round((d5-d2)/d2*100,2)}\% & :={round((e5-e2)/e2*100,2)}\% & :={round((f5-f2)/f2*100,2)}\% & :={round((g5-g2)/g2*100,2)}\% \\
  \midrule
  @ $\mathcal{L}_{CIoU}$ & 51.08 & 79.52 & 70.07 & 55.12 & 44.04 & 13.10 \\
  @ rela. improv.  & :={round((b7-b2)/b2*100,2)}\% & :={round((c7-c2)/c2*100,2)}\%
  & :={round((d7-d2)/d2*100,2)}\% & :={round((e7-e2)/e2*100,2)}\% & :={round((f7-f2)/f2*100,2)}\% & :={round((g7-g2)/g2*100,2)}\% \\
  \midrule
     @ $\mathcal{L}_{SCA}$ & {\fontseries{b}\selectfont}\STcopy{>}{51.84} & {\fontseries{b}\selectfont}\STcopy{>}{80.21} & {\fontseries{b}\selectfont}\STcopy{>}{70.91} & {\fontseries{b}\selectfont}\STcopy{>}{56.18} & {\fontseries{b}\selectfont}\STcopy{>}{45.14} & {\fontseries{b}\selectfont}\STcopy{>}{13.77} \\
 @rela. improv.& \textbf{:={round((b9-b2)/b2*100,2)}}\% & \textbf{:={round((c9-c2)/c2*100,2)}}\% &
 \textbf{:={round((d9-d2)/d2*100,2)}}\% & \textbf{:={round((e9-e2)/e2*100,2)}}\% & \textbf{:={round((f9-f2)/f2*100,2)}}\% & \textbf{:={round((g9-g2)/g2*100,2)}}\%\\
  \bottomrule
  \end{spreadtab}
  }
    \caption{Comparison between the performance of Faster-RCNN trained using
  $\mathcal{L}_{IoU}$, $\mathcal{L}_{GIoU}$, $\mathcal{L}_{DIoU}$, $\mathcal{L}_{CIoU}$, and $\mathcal{L}_{SCA}$ losses on the PASCAL VOC 2007 \textit{test} set.}
  \label{faster_voc}
 \end{table}
 
 \begin{table*}[h!]

\centering
\begin{spreadtab}{{tabular}{c c c c c c c c c c }}
 \toprule
 @Loss & @mAP & @AP$_{50}$ & @ AP$_{75}$ & @ AP$_{s}$ & @AP$_{m}$ & @AP$_{l}$ & @AP$_{r}$ & @AP$_{c}$ & @AP$_{f}$ \\
 \midrule\midrule
 @ $\mathcal{L}_{IoU}$ & 16.7 & 29.4 & 16.2 & 13.1 & 22.6 & 26.3 & 3.8 & 14.2 & 25.1 \\
 \midrule
 @ $\mathcal{L}_{GIoU}$ & :={round(17.00009/1.0, 3)} & 29.1 & 17.2 & round(13.0, 3) & 23.3 & 26.7 & {4.4} & 14.4 & 25.3\\
 @relative improv.(\%)  & :={round((b3-b2)/b2*100,2)}\% & :={round((c3-c2)/c2*100,2)}\%
 & :={round((d3-d2)/d2*100,2)}\% & :={round((e3-e2)/e2*100,2)}\% & :={round((f3-f2)/f2*100,2)}\% & :={round((g3-g2)/g2*100,2)}\% & :={round((h3-h2)/h2*100,2)}\% & :={round((i3-i2)/i2*100,2)}\% & :={round((j3-j2)/j2*100,2)}\%\\
 \midrule
 @ $\mathcal{L}_{DIoU}$ &  16.8 & 29.5 & 16.6 & 13.3 & round(23.2, 3) & 26.2 & 3.3 & 14.5 & 25.2 \\		
 @relative improv.(\%)  & :={round((b5-b2)/b2*100,2)}\% & :={round((c5-c2)/c2*100,2)}\%
 & :={round((d5-d2)/d2*100,2)}\% & :={round((e5-e2)/e2*100,2)}\% & :={round((f5-f2)/f2*100,2)}\% & :={round((g5-g2)/g2*100,2)}\% & :={round((h5-h2)/h2*100,2)}\% & :={round((i5-i2)/i2*100,2)}\% & :={round((j5-j2)/j2*100,2)}\%\\
 \midrule
 @ $\mathcal{L}_{CIoU}$ & 16.7 & 29.5 & 16.5 & 13.2 & 23.2 & 26.1 & 3.3 & 14.3 & 25.3 \\
 @relative improv.(\%)  & :={round((b7-b2)/b2*100,2)}\% & :={round((c7-c2)/c2*100,2)}\%
 & :={round((d7-d2)/d2*100,2)}\% & :={round((e7-e2)/e2*100,2)}\% & :={round((f7-f2)/f2*100,2)}\% & :={round((g7-g2)/g2*100,2)}\% & :={round((h7-h2)/h2*100,2)}\% & :={round((i7-i2)/i2*100,2)}\% & :={round((j7-j2)/j2*100,2)}\%\\
 \midrule
  @ $\mathcal{L}_{SCA}$ & {\fontseries{b}\selectfont}\STcopy{>}{17.7}  & {\fontseries{b}\selectfont}\STcopy{>}{30.4} & {\fontseries{b}\selectfont}\STcopy{>}{18.} & {\fontseries{b}\selectfont}\STcopy{>}{13.5} & {\fontseries{b}\selectfont}\STcopy{>}{24.1} & {\fontseries{b}\selectfont}\STcopy{>}{28.3}
  & {\fontseries{b}\selectfont}\STcopy{>}{4.9} & {\fontseries{b}\selectfont}\STcopy{>}{15.4}
  & {\fontseries{b}\selectfont}\STcopy{>}{25.9}\\
 @relative improv.(\%) & \textbf{:={round((b9-b2)/b2*100,2)}}\% & \textbf{:={round((c9-c2)/c2*100,2)}}\% &
 \textbf{:={round((d9-d2)/d2*100,2)}}\% & \textbf{:={round((e9-e2)/e2*100,2)}}\% & \textbf{:={round((f9-f2)/f2*100,2)}}\% & \textbf{:={round((g9-g2)/g2*100,2)}}\% & \textbf{:={round((h9-h2)/h2*100,2)}}\% & \textbf{:={round((i9-i2)/i2*100,2)}}\%
 & \textbf{:={round((j9-j2)/j2*100,2)}}\%\\
 \bottomrule
 \end{spreadtab}
  \caption{Comparison between the performance of Faster R-CNN trained using
 $\mathcal{L}_{IoU}$, $\mathcal{L}_{GIoU}$, $\mathcal{L}_{DIoU}$, $\mathcal{L}_{CIoU}$, and $\mathcal{L}_{SCA}$ losses on the LVIS1.0 \textit{val} set.}
 \label{faster_lvis}
\end{table*}
 
 \paragraph{Faster R-CNN on LVIS}
Similarly, we train Faster R-CNN on the LVIS1.0 dataset using the aforementioned bounding box regression losses for 6 epochs, and training images are resized such that its shorter edge is 800 pixels while the longer edge is no more than 1333. The results are shown in Table~\ref{faster_lvis}.
We can observe that the SCA loss surpasses existing losses consistently in terms of mAP and $AP_{75}$ compared with other IoU-based losses. To be more specific, SCA achieves 1.0 point mAP and 1.8 point $AP_{75}$ higher than IoU loss, respectively. The superiority of SCA loss is more pronounced at high accuracy levels,
which reach 11\% relative improvement at AP75. The improvements of SCA loss mostly come all frequent categories, $\emph{i.e.}$, it improves 1.1 point $AP_r$, 1.2 points $AP_c$, and 1.1 points $AP_f$ comparing with IoU loss, respectively.

\begin{table}[h]
	\centering
	
	\resizebox{.98\columnwidth}{!}{
	\begin{tabular}{ccccc|c}
			\toprule
			&IoU & Center Distance & SO & Corner Distance & mAP \\ \midrule
			(a) &  & \checkmark & & & 19.17\\ 
			(b) & & & & \checkmark & \textbf{52.64}\\ 
			\midrule
	        (c) &	\checkmark& & &  & 52.28 \\
			(d) & \checkmark& \checkmark& &  & 52.65\\
			(e) & \checkmark& & & \checkmark & 52.97  \\
			(f) & &  & \checkmark &  &  53.07 \\
			(g) & &  & \checkmark& \checkmark & \textbf{53.45} \\ 
			\bottomrule
	\end{tabular}
	}
	\caption{The contributions of the proposed components on PASCAL VOC test set with SSD.}
	\label{tab:loss_item}
	
\end{table}

\subsection{Ablation Study of Each Loss Item}
In this section, we conduct the experiments on PASCAL VOC with SSD to clarify the contributions of the proposed Corner Distance~(CD) loss, Side Overlap~(SO) loss, and the results are shown in Table~\ref{tab:loss_item}. Firstly, Side Overlap and Corner Distance can be separate as a loss while Center Distance cannot~(see (a), (b), and  (f)). In the same time, Corner Distance with IoU loss can be more powerful than Center Distance with IoU loss (DIoU loss), comparing the (d) and (e). Furthermore, CD loss can achieve the better performance than IoU loss (see (b) and (c)). Secondly, SO loss can bring substantial improvement than IoU loss~(+0.8 mAP, (c) and (f)). Finally, the overall performance is +1.2mAP than the IoU, which shows the superiority of our SCALoss.

\subsection{Ablation Study of Weight Factor $\alpha$}
In this section, we study the weight factor $\alpha$ for $\mathcal{L}_{SO}$ and $\mathcal{L}_{CD}$ in Eq.~(\ref{final_loss}). We use the SSD detection framework and the PASCAL VOC dataset to conduct experiments. The input image is resized to $300\times300$. We replace the $\ell_1$-smooth with our SCALoss and use different $\alpha = \{0.2, 0.3, ..., 0.7\}$ to show the importance of this parameter. 
The results have been shown in Table~\ref{alpha_study}.

For different models and datasets, we can make efforts to search an optimal $\alpha$ for better performance. However, for simplicity and saving computational resources, we choose $\alpha = 0.5$ for all our settings.

\begin{table}[h!]

  \centering
\resizebox{.98\columnwidth}{!}{
   \begin{tabular}{c c c c c c c} 
   \toprule
   $\alpha$ & mAP & AP$_{50}$ & AP$_{65}$ & AP$_{75}$ & AP$_{80}$ & AP$_{90}$ \\
   \midrule\midrule
   0.2 & 53.23 & \textbf{79.05} & 69.29 & 57.01 & 48.91 & 21.56 \\ 
   0.3 & 53.26 & 78.38 & \textbf{69.35} & 57.63 & 49.06 & 21.24 \\
   0.4 & 53.34 & 78.59 & 69.23 & \textbf{58.01} & \textbf{49.25} & 21.54 \\
   0.5 & \textbf{53.45} & 78.82 & 69.17 & 57.22 & 48.49 & \textbf{22.67} \\
   0.6 & 53.33 & 78.51 & 69.26 & 57.33 & 48.02 & 22.81 \\
   0.7 & 53.23 & 78.39 & 68.93 & 57.21 & 48.77 & 21.87 \\
   \bottomrule
   \end{tabular}
  }
   \caption{Study for different weight factor $\alpha$ for SSD trained using $\mathcal{L}_{SCA}$ on the PASCAL VOC 2007 \textit{test} set.}
   \label{alpha_study}
\end{table}

\section{Conclusions}

In this paper, we propose Side and Corner Aligned Loss~(SCALoss) for bounding box regression. SCALoss consists of Side Overlap and Corner Distance, which takes bounding box side and corner points into account. Combine the advantage of these two parts, SCALoss not only produces more penalty for low overlapping boxes and focuses more on hard samples but also speeds up the model convergence. In a result, SCALoss can serve as a more comprehensive measure than $\ell_n$ loss and IoU-based loss. Experiments on COCO, PASCAL VOC, and LVIS benchmarks show that SCALoss can bring consistent improvement and outperform $\ell_n$ loss and IoU based loss with popular object detectors, such as YOLOV3, SSD, and Faster R-CNN. 

In the future, we plan to investigate the feasibility of deriving an extension for SCALoss in the case of 3D object detection. This extension is promising to improve the performance of 3D object detection. 

\section*{Acknowledgments}
This work was supported in part by The National Key Research and Development Program of China (Grant Nos: 2018AAA0101400), in part by The National Nature Science Foundation of China (Grant Nos: 62036009, 61936006), in part by Innovation Capability Support Program of Shaanxi (Program No. 2021TD-05).

\bibliography{aaai22}

\end{document}